\documentclass[10pt]{article}

\usepackage[utf8]{inputenc}
\usepackage[T1]{fontenc}
\usepackage{times}
\usepackage{lipsum}
\usepackage{graphicx}
\usepackage{amsmath,amssymb}
\usepackage{hyperref}
\usepackage{cite}
\usepackage{geometry}
\usepackage{multirow}
\usepackage{longtable}
\usepackage[most]{tcolorbox}
\usepackage{array}
\usepackage{ragged2e}
\usepackage{booktabs}
\newcolumntype{P}[1]{>{\RaggedRight\arraybackslash}p{#1}}
\usepackage{subcaption}
\usepackage{floatrow}
\usepackage{blindtext}
\usepackage{comment}
\usepackage[font=small,labelfont=bf,tableposition=top]{caption}
\usepackage{listings}
\providecommand{\keywords}[1]{\noindent\textbf{Keywords—} #1}

\title{Implicature in Interaction: Implicature-Aware Prompting Improves User Evaluations of LLM Responses}
\author{%
  Asutosh Hota \quad Jussi P. P. Jokinen\\
  \small \texttt{asutosh.jyu.hota@jyu.fi} \quad \texttt{jussi.p.p.jokinen@jyu.fi}\\
  \small Jyvaskylan yliopisto, Jyväskylä, Central Finland, FI \\
}

\date{} 

\begin{document}
\maketitle
\vspace{-2em}
\begin{abstract}
The rapid advancement of Large Language Models (LLMs) is positioning language at the core of human--computer interaction (HCI). This shift increases the importance of pragmatic phenomena such as implicature, where users convey meaning beyond what is explicitly stated. We examine whether making implied communicative intent explicit during interaction improves human evaluation of LLM responses. To do so, we introduce an HCI-oriented operational taxonomy of three implicature classes in human--LLM interaction: information-seeking, direction-seeking, and expressive. We then report three experiments. Experiment~1 examines whether the proposed three-class implicature distinction is interpretable and can be applied with reasonable consistency on an authored prompt set, and benchmarks multiple LLMs against a human interpretation baseline. Experiment~2 tests whether an implicature-aware prompting intervention improves perceived relevance and quality of model responses. Experiment~3 uses a targeted forced-choice follow-up with a new participant sample to examine whether selected contrasts from Experiment~2 also translate into direct pairwise preference. Results show that larger models align more closely with human interpretations, and that implicature-aware prompting improves perceived relevance and quality across the evaluated response set. In the targeted follow-up, participants preferred implicature-aware responses more often than baseline responses. These findings suggest that surfacing implied communicative intent can improve the perceived quality of language-based interaction and highlight the value of integrating linguistic theory into HCI research and design.
\end{abstract}

\begin{figure}[ht]
    \centering
    \includegraphics[width=\textwidth]{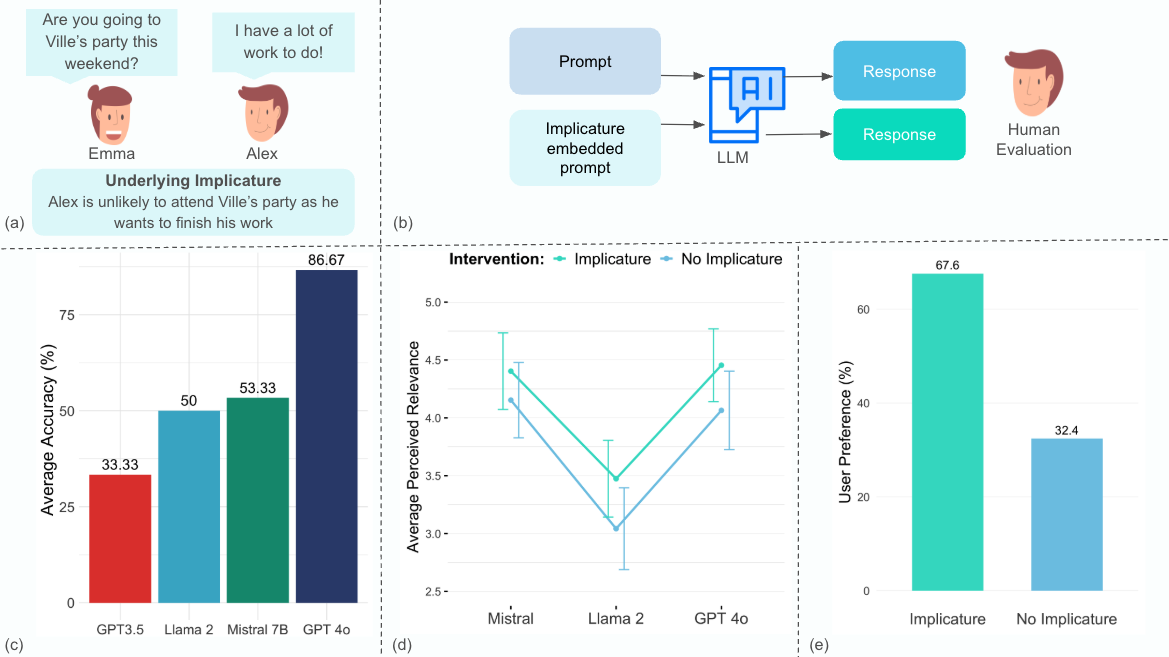}
    \caption{Overview of the study. (a) Example of conversational implicature. (b) Study logic comparing baseline and implicature-aware prompting conditions. (c) Experiment~1 benchmarks model interpretations against a human interpretation baseline. (d) Experiment~2 evaluates perceived relevance and quality of generated responses. (e) Experiment~3 provides a targeted forced-choice follow-up on selected response pairs, showing that implicature-aware responses were preferred more often than baseline responses.}
    \label{fig:fig1}
\end{figure}

\section{Introduction}
Human--computer interaction (HCI) is entering a linguistic turn.
As Large Language Models (LLMs) are embedded into digital assistants, chatbots, productivity tools, and social robots, language is increasingly becoming the dominant interface through which users engage with computational systems \cite{pang2025understanding}.
Unlike earlier eras of HCI that emphasized graphical or direct-manipulation interfaces \cite{shneiderman1982future}, contemporary interaction increasingly depends on how well systems can interpret and respond to the subtleties of human communication \cite{sun2024generative}.
While LLMs have made notable progress in language generation, many still struggle with pragmatic aspects of language use and often default to literal interpretations when meaning depends on context \cite{grice1975logic, huang2017oxford, sravanthi2024pub, ruis2023goldilocks}.
This makes linguistic theory, including pragmatics and discourse, increasingly relevant to the future of HCI research.

A persistent challenge lies in interpreting meaning beyond the literal surface of words.
In everyday conversation, people routinely rely on \textit{conversational implicature}, where intended meaning must be inferred from context, shared assumptions, and communicative norms.
For example, replying ``I have a lot of work to do'' to an invitation can imply refusal without stating it explicitly.
Such indirect cues are central to how people express requests, refusals, attitudes, and social nuance.
When interaction systems fail to recognize these cues, users may experience dialogue as rigid, awkward, or unintelligent, which can reduce trust and weaken the sense of collaboration with the system \cite{shin2021effects, webson2022prompt}.
For language-based interaction, implicature is therefore not a marginal phenomenon but a practically important aspect of how users communicate with AI systems.

Although LLMs have achieved impressive fluency, their pragmatic competence remains uneven.
Some state-of-the-art models approach human-level performance on selected implicature benchmarks \cite{bojic2025does, yue2024large}, but smaller or open-source models often rely on literal interpretations or shallow heuristics \cite{ruis2023goldilocks, cho-ismkim99-skku-edu-2024-pragmatic}.
Moreover, because the reasoning behind LLM outputs is often opaque \cite{liu2023trainingsociallyalignedlanguage}, it remains difficult to diagnose pragmatic failures or translate linguistic theory into actionable HCI design guidance.
From an HCI perspective, this creates two related needs: first, a lightweight and operational way to characterize implicature in human--LLM interaction, and second, user-facing evidence about whether making implied intent explicit actually improves interaction quality.

This paper treats implicature as a concrete HCI test case for linking pragmatic theory to the design and evaluation of human--LLM interaction.
We operationalize three implicature classes that are common in language-based interaction settings: \textit{information-seeking} (indirect requests for knowledge), \textit{direction-seeking} (indirect requests for guidance or action), and \textit{expressive} (utterances conveying affect or evaluation).
Our goal is not to propose an exhaustive linguistic taxonomy, but to develop a pragmatic, HCI-oriented categorization with explicit conversational features that can support stimulus design, benchmarking, and user evaluation.

We report three experiments that use this operationalization to study implicature in human--LLM interaction. Experiment~1 asks whether humans can apply the rubric consistently and how closely LLM interpretations align with a human interpretation baseline. Experiment~2 examines whether making communicative intent explicit through an implicature-aware prompting intervention improves users' evaluations of response relevance and quality. Because Experiment~2 uses an incomplete between-subjects rating design and does not present both response variants side by side, Experiment~3 serves as a targeted forced-choice follow-up with a new participant sample to examine whether selected contrasts also translate into direct pairwise preference. Figure~\ref{fig:fig1} summarizes the study design and main findings. Across experiments, larger models align more closely with human interpretations, and responses generated with implicature-aware prompting are rated higher on relevance and quality than baseline responses. In the targeted follow-up, participants also preferred implicature-aware responses more often than baseline responses. Because our manipulation uses structured prompting, including system-level guidance, we interpret the intervention as an HCI-relevant \textit{prompting-based method for surfacing implied intent} rather than as a pure measure of intrinsic model implicature competence.

Our contributions are threefold:
\begin{enumerate}
    \item We introduce an HCI-oriented operational taxonomy of implicature classes with conversational features that supports systematic stimulus design and evaluation.
    \item We provide a benchmark of multiple LLMs against a human interpretation baseline for implicature interpretation on an authored stimulus set.
    \item We present user-study evidence that implicature-aware prompting improves user evaluations of LLM responses, specifically perceived relevance and perceived quality, along with increased preference in a targeted direct-comparison follow-up.
\end{enumerate}

\section{Background}

\subsection{Implicatures in Pragmatics and Linguistics}

\begin{table}[t]
  \caption{Major themes, learnings, and open challenges in implicature research.}
  \label{tab:implicature_summary}
  \begin{tabular}{p{0.26\linewidth} p{0.34\linewidth} p{0.34\linewidth}}
    \toprule
    \textbf{Theme} & \textbf{Key Learnings} & \textbf{Challenges} \\
    \midrule
    How implicatures are generated & 
    Gricean accounts compute implicatures post-semantically \cite{recanati1989pragmatics}, while grammatical accounts generate them locally in syntax \cite{chierchia2004scalar,chierchia2012grammatical}. &
    Clearer criteria are needed to avoid predicting implicatures that listeners do not actually infer. \\

    Embedded implicatures & 
    Implicatures can arise inside quantifiers or conditionals (e.g., ``Exactly one student solved some of the problems'') \cite{sauerland2012computation,chierchia2012grammatical}. &
    Psycholinguistic evidence is limited on whether people compute such inferences in real time. \\

    Plural/quantity implicatures & 
    Plural phrases often imply ``more than one'' but this varies with logical context \cite{spector2007aspects}. &
    Formal models must better capture context-sensitive variability. \\

    Context dependence of meaning & 
    Even basic meanings (e.g., ``John's book'') depend heavily on context \cite{recanati1989pragmatics,recanati2003embedded}. &
    Tools are needed to separate sentence meaning from pragmatic enrichment. \\

    Real-time processing & 
    Existing theories largely ignore processing effort or timing \cite{recanati1989pragmatics,chierchia2012grammatical}. &
    More cognitive studies are needed to test when and how implicatures are formed during conversation. \\
    \bottomrule
  \end{tabular}
\end{table}

Implicatures are a core mechanism of everyday communication in which speakers convey meaning indirectly and listeners infer that meaning from context, shared assumptions, and conversational norms. A reply such as ``It's getting pretty late'' can function as a polite refusal without stating refusal explicitly. Because implicatures depend on context and assumptions about cooperative behaviour, they are a useful test case for any theory or system that aims to model communicative intent.

A central and well-studied case is scalar implicature \cite{chierchia2012scalar}, where a weaker term such as ``some'' is often interpreted as implying the negation of a stronger alternative such as ``all.'' Theoretical accounts differ in where this inference originates. Gricean approaches treat implicature as a post-semantic inference derived through pragmatic reasoning guided by conversational maxims \cite{recanati1989pragmatics,grice1975logic}. Grammatical approaches treat at least some implicatures as part of the compositional system and allow them to be generated locally \cite{chierchia2004scalar,chierchia2012grammatical}. Despite these differences, both traditions converge on an important point: implicature is systematic, but not automatic. The same surface form can support multiple intended meanings, and listeners do not draw the same inference in every context.

This variability is precisely what makes implicature difficult to model. Some theories risk overpredicting implicatures across contexts, while others risk missing inferences that people do make in practice. Work on context dependence \cite{recanati1989pragmatics,recanati2003embedded} further shows that even apparently simple expressions can be pragmatically enriched in different ways depending on shared knowledge and situational cues. Work on processing \cite{recanati1989pragmatics,chierchia2012grammatical} suggests that implicature formation may also depend on timing, salience, and cognitive effort. Table~\ref{tab:implicature_summary} summarises several of these themes and their open challenges.

For HCI, the main implication is that indirect meaning cannot be handled reliably by surface matching alone. If interaction systems are to respond naturally to users, they must accommodate the fact that meaning is often conveyed through implication rather than direct statement. At the same time, HCI does not necessarily require a comprehensive linguistic taxonomy of all implicature phenomena. What it needs is a practical operationalization that captures common interaction-facing intents and can be used to support both stimulus design and evaluation.

\subsection{Implicatures in Human-Computer Interaction (HCI)}

\begin{table}[t]
  \caption{Themes, learnings, and design challenges of implicature in HCI.}
  \label{tab:hci_implicature_summary}
  \begin{tabular}{p{0.26\linewidth} p{0.34\linewidth} p{0.34\linewidth}}
    \toprule
    \textbf{Theme} & \textbf{Key Learnings} & \textbf{Challenges} \\
    \midrule
    Annotated datasets for implicature & Datasets like PUB, CIRCA, and those from Ruis et al. \cite{george2020conversational,anuranjana2024survey,ruis2023goldilocks} help model pragmatic meaning beyond literal content. & High annotation cost and limited domain diversity limit generalization. \\
    Model performance and explanation & Models like GPT-4 approach human-level accuracy with tailored tuning \cite{ruis2023goldilocks,grice1991studies}. & Many models struggle with explaining implicatures logically \cite{yue2024large}. \\
    Collaborative and cooperative implicature & AI that uses implicature can enhance perceived humanness and coordination \cite{liang2019implicit}. & Requires shared context and alignment on conversational goals. \\
    Implicitness in UI and system behavior & Implicit cues span attention, intent, and utterance meaning \cite{serim2019explicating}. & Disambiguating implicit interaction types is critical for design consistency. \\
    User perception and agent framing & Implicature interpretation changes based on whether users see AI as autonomous or tool-like \cite{nishihata2023human}. & Systems should signal autonomy and transparency to guide expectations. \\
    Cognitive architecture for intent modeling & Combining LLMs with symbolic models improves intent recognition \cite{iida2024integrating}. & Integration is computationally expensive and design-intensive. \\
    Visual and multimodal implicature & Layout and gesture carry implicature-like inferences \cite{kehler2000cognitive,oberlander1995grice}. & Multimodal systems must consider indirect meanings from visual design. \\
    Maxim adherence in Turing-style evaluation & Violating Gricean norms reveals artificiality in chatbots \cite{saygin2002pragmatics}. & Adherence to conversational maxims is crucial for believable interaction. \\
    \bottomrule
  \end{tabular}
\end{table}

Linguistic theory has long informed HCI research, especially work that treats interaction as situated communication rather than mere information transfer. Pragmatics \cite{saygin2002pragmatics,recanati1989pragmatics} has shaped how researchers think about relevance, informativeness, and clarity in conversational systems, while speech act theory \cite{serim2019explicating,searle1975indirect} has been used to describe how utterances function as requests, commitments, and expressions. Research on implicitness and multimodality further shows that meaning in interaction is often conveyed through cues that are not explicitly stated in text, including visual layout and gesture \cite{serim2019explicating,kehler2000cognitive,oberlander1995grice}. Within this broader tradition, implicature is a particularly useful focal point because it provides a concrete and testable instance of indirect meaning, while also bearing directly on how cooperative or intelligent an interaction system appears \cite{saygin2002pragmatics}.

Recent work in HCI and adjacent NLP has also produced datasets and benchmarks for evaluating implicature comprehension. George and Mamidi \cite{george2020conversational} collected implicature-rich dialogues from movies, while PUB \cite{anuranjana2024survey} and its extensions \cite{sravanthi2024pub,ruis2023goldilocks} evaluate models on pragmatically implied meaning. These resources have been valuable for controlled comparison across models and training regimes. However, they are often narrow in domain coverage, costly to annotate, and oriented more toward benchmark performance than toward the needs of interaction design. As a result, they do not directly provide a lightweight, HCI-oriented categorization that can be used to generate stimuli, construct prompting interventions, and evaluate common user-facing interaction intents.

A related issue concerns model behaviour and explainability. Even when models arrive at the correct answer on implicature tasks, they may not provide reliable or coherent explanations of how that answer was reached. Yue et al. \cite{yue2024large} show that LLMs can struggle to justify implicature-based choices, while Cong \cite{cong2024manner} reports persistent failures on manner implicatures, which depend on how something is said rather than only what is said. These findings matter for HCI because interaction quality depends not only on whether a system succeeds on a benchmark, but also on whether its behaviour appears accountable, transparent, and trustworthy to users.

Beyond benchmark performance, implicature also matters in interactive and social settings. In \textit{Hanabi}, Liang et al. \cite{liang2019implicit} show that agents using implicature-like strategies can improve coordination and appear more human-like. Other work shows that implicit meaning extends beyond language alone: layout, gesture, and other interface cues can create implicature-like inferences \cite{kehler2000cognitive,oberlander1995grice}. User framing also shapes interpretation. Nishihata et al. \cite{nishihata2023human} show that indirect cues are judged differently depending on whether an agent is framed as autonomous or tool-like. Hybrid approaches that combine LLMs with cognitive or symbolic models can improve intent modelling \cite{iida2024integrating}, but such approaches are often computationally expensive and design-intensive. More broadly, work on conversational maxims suggests that violations of relevance or quantity can make systems feel artificial or uncooperative \cite{saygin2002pragmatics}. Table~\ref{tab:hci_implicature_summary} summarises these themes and design challenges.

Taken together, prior HCI and NLP work leaves two gaps that motivate our study. First, although existing datasets and benchmarks enable controlled testing, they do not provide a lightweight, interaction-oriented operationalization that supports stimulus design, prompt construction, and evaluation across common user intents. Second, benchmark accuracy alone does not establish whether implicature-sensitive handling improves perceived interaction quality in user-facing settings \cite{shin2021effects,webson2022prompt}. Our work addresses these gaps by operationalising three implicature classes that map to common communicative purposes and by evaluating whether an implicature-aware prompting intervention is associated with higher perceived relevance, higher perceived quality, and greater user preference.

\subsection{Background Summary}

Prior work in linguistics shows that implicature is systematic but highly context-sensitive, making it a demanding but informative target for evaluating pragmatic understanding. Prior work in HCI and NLP has developed useful datasets, benchmarks, and design insights, but it has tended to prioritise linguistic coverage and model comparison over lightweight, interaction-oriented operational categories. At the same time, user-facing research suggests that pragmatic misalignment affects perceptions such as trust, naturalness, and cooperativeness, and that indirect meaning can be conveyed through multiple modalities and framings.

Across these strands, Grice's cooperative principle \cite{grice1975logic,grice1991studies} and Searle's speech act taxonomy \cite{searle1975indirect} provide useful conceptual anchors for operationalisation. Users routinely rely on indirect meaning to seek information, request guidance, or express evaluation. Building on this foundation, our study operationalises three implicature classes---information-seeking, direction-seeking, and expressive---and evaluates them across three experiments using both interpretation outcomes and user-perceived relevance, quality, and preference.

Crucially, our paper asks a different question from most benchmark work. Rather than asking only whether models can infer implicature, we ask whether making implied communicative intent explicit through prompting improves user-evaluated response quality in human--LLM interaction.

\section{Method}

This study comprises three connected experiments that examine implicature in human--LLM interaction from complementary perspectives. Experiment~1 examines whether the proposed three-class implicature distinction can be applied with reasonable consistency on an authored prompt set and evaluates how closely model interpretations align with a human interpretation baseline. Experiment~2 serves as the main user study and tests whether making communicative intent explicit through an implicature-aware prompting intervention improves perceived relevance and overall quality of LLM responses. Because Experiment~2 uses an incomplete between-subjects rating design and does not present both response variants side by side, Experiment~3 uses a new participant sample and a forced-choice paradigm on a targeted subset of response pairs to examine whether selected contrasts from Experiment~2 also translate into direct pairwise preference.

\subsection{Taxonomy generation and operationalization}
\label{sec:taxonomy_generation}

Our goal in developing the taxonomy was practical rather than exhaustive. We sought a lightweight, HCI-oriented categorization that could support stimulus design and evaluation for common human--LLM interaction intents while remaining based on pragmatic theory. Accordingly, we prioritized classes that (i) occur frequently in everyday prompts to conversational systems, (ii) differ in what a cooperative response should do, and (iii) can be applied with explicit and reasonably stable labeling rules.

We drew on two theoretical anchors, summarized in Table~\ref{tab:taxonomy_overview}. First, Grice's cooperative principle and conversational maxims provide an account of why users often imply communicative intent rather than stating it directly \cite{grice1975logic,grice1991studies}. Second, speech act theory helps characterize what utterances are doing in interaction, including information requests, directives, and expressions of affect \cite{searle1975indirect}. Rather than attempting to cover all implicature types discussed in linguistic theory, we used these perspectives to derive a small set of interaction-relevant categories that capture different first-turn response obligations.

\begin{table}[t]
\small
\caption{Implicature taxonomy overview. Each class is based on Gricean pragmatics \cite{grice1975logic,grice1991studies} and Searle's speech acts \cite{searle1975indirect}, and summarized as a compact rubric for labeling prompts by primary implied intent.}
\label{tab:taxonomy_overview}
\begin{tabular}{p{0.12\linewidth} p{0.20\linewidth} p{0.26\linewidth} p{0.18\linewidth} p{0.15\linewidth}}
\toprule
\textbf{Class} & \textbf{Grounding} & \textbf{Conversational features} & \textbf{Common cues} & \textbf{Exclude if} \\
\midrule
Information-seeking &
Grice: Relation (sometimes Quantity). Searle: assertives/questions. &
Indirect questions that imply a need for clarification or planning (e.g., ``Will it rain tomorrow?'' implies planning for the day). &
Curiosity, confusion, uncertainty, ``can you explain'', ``what does X mean'' and indirect statements that imply a question. &
User mainly asks for next steps or expresses affect as the main goal. \\
\midrule
Direction-seeking &
Grice: flouting of Relation or Manner. Searle: directives (indirect speech acts). &
Declarative statements that function as veiled requests for guidance or help (e.g., ``I don't know how to start this assignment''). &
Problem state, being stuck, weighing alternatives, ``what should I do'' and indirect help-seeking. &
User mainly wants an explanation without requesting action guidance. \\
\midrule
Expressive &
Grice: flouting of Quantity or Quality (figurative, evaluative). Searle: expressives. &
Utterances that convey emotions or attitudes rather than facts (e.g., ``That was surprisingly fun'' signals a positive evaluation). &
Strong sentiment, celebration, frustration, relief, gratitude and evaluative statements without a request. &
User mainly requests factual information or procedural guidance as the primary goal. \\
\bottomrule
\end{tabular}
\end{table}

The taxonomy was developed iteratively. We began with speech-act-like functions that frequently appear in prompts to conversational systems: requests for information, requests for guidance, and expressions of affect or evaluation. We then examined examples from prior implicature-focused resources in HCI and NLP \cite{george2020conversational,anuranjana2024survey,ruis2023goldilocks} and refined the class definitions based on two criteria: whether prompts within a candidate class shared a stable implied intent, and whether different classes implied meaningfully different response obligations. This process yielded three classes---information-seeking, direction-seeking, and expressive---that are theory-consistent while remaining operational for HCI-oriented evaluation.

We define each class by the implied intent that a cooperative assistant should prioritize in its first response. Information-seeking prompts imply a question or a desire for explanation, so the appropriate response is primarily informative or clarificatory. Direction-seeking prompts imply a request for guidance, recommendation, or next steps, so the appropriate response is primarily action-oriented. Expressive prompts primarily convey affect, attitude, or evaluation and therefore imply a social response obligation such as acknowledgement, affirmation, or empathy; information or guidance may be added, but are secondary to recognizing the user's stance.

\subsubsection{Labeling rubric}
\label{sec:taxonomy_rubric}

To assign prompts to classes, we used a rubric centered on \emph{primary implied intent} and \emph{first-turn response obligation}. If a prompt contained an explicit request for explanation or factual clarification, it was labeled \textit{Information-seeking}. If it explicitly requested advice, actions, or next steps, it was labeled \textit{Direction-seeking}. If it primarily expressed an emotion, attitude, or evaluation without centering a request, it was labeled \textit{Expressive}.

When prompts did not contain an explicit request, we labeled them according to what a cooperative assistant should address first. Indirect statements implying a need for explanation (e.g., ``I keep hearing about X and I do not get it'') were labeled \textit{Information-seeking}. Indirect statements implying a need for guidance (e.g., ``My laptop is slow and keeps crashing'') were labeled \textit{Direction-seeking}. Prompts whose primary function was social sharing, affect, or evaluation (e.g., ``I feel so relieved today'') were labeled \textit{Expressive}.

For mixed-intent prompts, we resolved ambiguity using first-turn response obligation. If empathy or acknowledgement should come before advice or factual explanation, the prompt was labeled \textit{Expressive}. If the first cooperative move should be to offer steps, options, or recommendations, the prompt was labeled \textit{Direction-seeking}. If the first cooperative move should be to clarify or explain, the prompt was labeled \textit{Information-seeking}. This rule was especially important for prompts that combined frustration with help-seeking, or evaluation with a partially implied question.

We also defined boundary-case rules to improve consistency. Rhetorical questions with strong affect (e.g., ``Why does this always happen to me?'') were labeled \textit{Expressive} unless the user clearly sought a technical explanation. Problem descriptions without an explicit question were labeled \textit{Direction-seeking} when they implied a desire to fix the problem. Opinions or evaluative statements about a topic were labeled \textit{Expressive} unless they clearly implied a request for information.

\subsection{Stimulus construction and response generation}
\label{sec:data_generation}

We authored 30 prompts to instantiate the three implicature classes, with 10 prompts per class. Prompts were designed to resemble realistic user utterances in human--LLM interaction and were written using the rubric described in Section~\ref{sec:taxonomy_rubric}. The same prompt set was reused across all three experiments so that differences between experiments would reflect the evaluation task and prompting condition rather than changes in stimuli.

Within each class, prompts were designed to share a similar primary implied intent and first-turn response obligation. We also included mixed-intent prompts, such as affect combined with help-seeking, and assigned a primary class using the boundary-case rules above. Because the prompts were authored using the rubric itself, the results from Experiment~1 should be interpreted as evidence of learnability and internal coherence on this stimulus set, rather than as external validation on naturally occurring prompts.

For response generation, each prompt was paired with two model-response conditions. Both conditions used the same shared base system setup to establish a helpful, context-aware assistant role.

In the \emph{implicature-aware} condition, models were given a brief definition of the relevant implicature class and its communicative goal, similar description which was provided to human participants in experiment 1. The implicature-aware condition did not provide the model with the full taxonomy rubric or detailed decision rules. Rather, it supplied a brief definition of the inferred implicature class so that the model could orient its response to the likely communicative goal of the prompt.
In the \emph{baseline} condition, no class-specific guidance was added. We therefore interpret this manipulation as an \emph{implicature-aware prompting intervention} that makes communicative intent explicit, rather than as a direct test of unaided implicature inference. All responses were generated in a zero-shot setup using standardized decoding settings across models.

\subsection{Participants}

Participants were recruited via Prolific, and all participants provided informed consent electronically (written) through the platform before taking part. The study did not collect additional personally identifying or sensitive personal data beyond the information required for participation and compensation. 
The study was conducted in accordance with the principles of the Declaration of Helsinki and in accordance with the research practices of the Cognitive Science research group, Faculty of Information Technology, University of Jyv\"askyl\"a, Finland. As this was a minimal-risk anonymous online study, no separate ethics permit number was issued under the applicable local institutional practice. Attention checks were included in all experiments, and incomplete responses or responses failing quality-control criteria were excluded prior to analysis.

For Experiment~1, 60 participants were recruited, and 51 were retained in the final analysis after sanity checks/complete submissions. Demographic data were available for 50 of these participants via Prolific records (32 female, 18 male; mean age = 35.8 years, SD = 10.4).
For Experiment~2, 90 participants were recruited via Prolific. Of these, 84 were retained in the final analysis after excluding responses based on attention/sanity checks and incomplete or invalid submissions. In the Prolific recruitment records, the recruited participant pool comprised 43 female, 44 male, and 3 participants who preferred not to say (mean age = 35.9 years, SD = 13.0). For Experiment~3, 30 participants were recruited via Prolific, and 29 were retained in the final analysis after excluding one response based on quality-control criteria. 
In the Prolific recruitment records, the recruited participant pool comprised 19 female and 11 male participants (mean age = 34.5 years, SD = 10.6).


\subsection{Models evaluated}

Experiment~1 benchmarked five LLMs spanning a range of scales and model families: \textbf{GPT-4o}, \textbf{GPT-4}, \textbf{GPT-35-turbo} (Azure designation for GPT-3.5), \textbf{Llama-2-7b-chat}, and \textbf{Mistral-7B-Instruct-v0.1}. These models were selected to compare high-capacity proprietary models with smaller or open-access alternatives that differ in scale, instruction tuning, and likely pragmatic sensitivity. All models were accessed via API and prompted in a zero-shot configuration using matched decoding settings (temperature = 0.5; maximum output length = 1000 tokens).

Experiments~2 and~3 used a closely related but not identical model set: \textbf{GPT-4o}, \textbf{GPT-35-turbo}, \textbf{Llama-2-7b-chat}, \textbf{Mistral-7B-Instruct-v0.1}, and \textbf{Phi-3-small}. The model sets differ by design. \textbf{GPT-4} was included in Experiment~1 as part of the initial interpretation benchmark, but was excluded from the later user-facing experiments because \textbf{GPT-4o} showed stronger performance in Experiment~1 and made GPT-4 redundant for subsequent response-generation comparisons. In contrast, \textbf{Phi-3-small} was introduced in Experiments~2 and~3 to preserve variation among smaller-capacity models in the user-facing evaluations. This substitution allowed us to retain diversity in model scale and capability while avoiding unnecessary duplication between two closely related high-capacity models.

Accordingly, Experiment~1 should be interpreted as a broad benchmark of model alignment with a human interpretation baseline, whereas Experiments~2 and~3 evaluate the effects of implicature-aware prompting on a user-facing response set chosen to balance model diversity, participant burden, and practical redundancy.
\subsection{Experiments}

The three experiments were designed to address complementary questions about implicature in human--LLM interaction. Experiment~1 establishes whether the proposed three-class implicature distinction is interpretable to independent human annotators and benchmarks model interpretations against a human interpretation baseline on the authored prompt set. Experiment~2 serves as the main user-facing study and tests whether making communicative intent explicit through implicature-aware prompting improves perceived relevance and perceived quality of LLM responses. Because Experiment~2 uses an incomplete between-subjects rating design and does not present both response variants side by side, Experiment~3 provides a targeted forced-choice follow-up with a new participant sample to examine whether selected contrasts from Experiment~2 also translate into direct pairwise preference. Together, the three studies move from interpretability of the operationalization, to user evaluation of generated responses, to direct comparison of selected response pairs. The full system prompts used in the study are provided in Appendix~\ref{appendix}, including Listing~\ref{lst:exp1-system-prompt} for the implicature classification prompt in Experiment~1 and Listing~\ref{lst:system-prompt} for the response-generation prompt used in Experiment~2.

\subsubsection{Experiment~1: Implicature interpretation benchmark}

Experiment~1 had two purposes: to evaluate whether the proposed three-class distinction was interpretable and could be applied with reasonable consistency on the authored prompt set, and to benchmark model interpretations against a human interpretation baseline.

The 30 prompts used in this experiment were authored using the taxonomy rubric described earlier. However, independent human annotators were not given the full rubric or decision procedure used to construct the prompts. Instead, they were provided only with brief definitions of the three classes and asked to assign each prompt to exactly one class. For each prompt, we computed the empirical distribution of human labels and the modal human label. The empirical distribution captures how the prompt was perceived by independent annotators, while the modal label provides a single reference label for prompt-level agreement comparison.

The same 30 prompts were then presented to each LLM, which was asked to assign one of the three classes. As with the human annotators, the task was framed at the level of class definitions rather than full rubric reproduction. Model performance was evaluated by comparing each model's predicted label against the human baseline derived from the annotator data. We report both agreement metrics among annotators and model alignment with the human interpretation baseline.

This experiment should not be interpreted as establishing a universal ground truth for implicature labeling or as externally validating the taxonomy on naturally occurring prompts. Rather, it tests whether prompts authored to instantiate the three classes are recognizably classifiable by independent annotators given class-level descriptions, and whether model interpretations align with that human baseline.

\subsubsection{Experiment~2: Perceived relevance and quality}
\label{sec:exp2}
Experiment~2 served as the main user-study evaluation of the prompting intervention. Participants rated model-generated responses on two dimensions: \emph{perceived relevance} (how well the response aligned with the implied intent of the prompt) and \emph{overall quality} (clarity, coherence, and usefulness). Both measures were collected on 5-point Likert scales.

The experiment used the same 30 prompts developed for Experiment~1. Each prompt had two response variants: one generated under the implicature-aware condition and one generated under the baseline condition. To keep the task manageable, participants did not view the full set of possible items. Instead, each participant rated a subset of 20 prompt--response items sampled from the larger stimulus pool. This produced an \emph{incomplete between-subjects design}: participants saw only one response at a time and did not directly compare the implicature-aware and baseline responses for the same prompt. 

Prompt--response items were presented in randomized order. Because participants were shown only a subset of items, Experiment~2 used an incomplete between-subjects design: participants did not directly compare the implicature-aware and baseline responses for the same prompt.
The sample shown to each participant included a mixture of prompt classes, prompting conditions, and model outputs. Attention-check items were interspersed to maintain data quality. This design allowed us to estimate condition differences through aggregate ratings across participants while keeping individual workload at a reasonable level.
Figure~\ref{fig:exp2-instruction} shows the instructions presented to participants in Experiment~2, and Figure~\ref{fig:exp2-example} shows an example prompt--response item from the task.

\subsubsection{Experiment~3: Targeted forced-choice follow-up}
\label{sec:exp3}

Experiment~3 used a two-alternative forced-choice design with a new participant sample. Its purpose was not to provide an independent estimate of overall preference across the full prompt set, but to serve as a focused follow-up to Experiment~2. Because Experiment~2 used an incomplete between-subjects rating design, participants in that study did not directly compare implicature-aware and baseline responses for the same prompt. Experiment~3 therefore examined whether selected contrasts from Experiment~2 also translated into direct pairwise preference when both alternatives were shown side by side.

The comparison items for Experiment~3 were selected post hoc from the Experiment~2 response pool. Specifically, we first identified prompt--response pairs for which the difference between the implicature-aware and baseline conditions exceeded 2 points on either perceived relevance or perceived quality at the individual rating level. From this filtered set, we randomly sampled 10 comparison pairs for inclusion in Experiment~3. Thus, the final item set was not sampled independently from the full prompt pool, but randomly chosen from a subset of higher-contrast cases identified after Experiment~2. All participants in Experiment~3 saw the same 10 selected pairs, presented in randomized order.

The value of this experiment is therefore narrower than that of Experiment~2: it tests whether selected contrasts visible in the rating data also translate into direct pairwise choice when both alternatives are shown together. Moreover, because the implicature-aware condition makes the likely communicative intent explicit through brief class-based guidance, the study evaluates the effect of pragmatic scaffolding rather than unaided implicature inference.
Figure~\ref{fig:exp3-instruction} shows the instructions presented to participants in Experiment~3, and Figure~\ref{fig:exp3-example} shows an example forced-choice trial.

\subsection{Data analysis}

For Experiment~1, we used multi-rater agreement statistics to examine whether the proposed three-class implicature distinction could be applied with reasonable consistency on the authored prompt set. Specifically, we report Fleiss' $\kappa$ as a chance-corrected measure of multi-rater agreement, average pairwise agreement, and per-prompt modal agreement. To benchmark model interpretations, we computed agreement with the modal human label for each prompt and additionally examined how closely model outputs tracked the empirical human label distributions.

For Experiment~2, we analysed perceived relevance and perceived quality using linear mixed-effects models with fixed effects for prompting condition (implicature-aware vs.\ baseline), model, and implicature class, including all interactions. To account for repeated observations and item-level clustering, we included random intercepts for participant and prompt--response pair. Type~III tests were computed using Satterthwaite's method. As descriptive follow-up analyses, we estimated marginal means for prompting condition within each implicature class and applied Holm correction for multiple comparisons. Because the design is incomplete and between-subjects at the item level, the Experiment~2 analysis should be interpreted as estimating condition differences across the evaluated response set rather than as a within-person paired comparison. Exact variance components and full fixed-effect estimates for the Experiment~2 mixed-effects models are reported in Appendix~A.

For Experiment~3, we summarized the proportion of valid choices in which the implicature-aware response was selected over the baseline response and tested whether that proportion exceeded chance (50\%) using an exact binomial test. As Experiment~3 used a targeted post hoc subset of comparison items selected from Experiment~2, these results are interpreted as evidence about the selected comparison set rather than as an independent estimate of preference across the entire prompt pool. Unless otherwise noted, statistical significance was defined as $p < .05$.

\section{Results}

\subsection{Experiment 1: Implicature interpretation benchmark}
\label{sec:exp1_results}

Experiment~1 examined two related questions: whether the proposed three-class implicature rubric was learnable and internally coherent on the authored prompt set, and how closely different LLMs aligned with a human interpretation baseline derived from independent annotators.

\subsubsection{Rubric learnability and internal coherence}

We first assessed whether independent annotators could apply the rubric with non-trivial consistency. Across 51 annotators and 30 prompts, inter-annotator agreement was moderate (Fleiss' $\kappa = 0.414$, $z = 114$, $p < .001$; Table~\ref{tab:iaa}). Mean pairwise agreement was $\bar{P} = 0.612$. At the prompt level, median modal agreement was 0.716, with values ranging from 0.431 to 1.000. Taken together, these results indicate that the rubric was learnable and usable on the authored stimulus set, while also showing that some prompts were more ambiguous than others.

The prompt-level agreement statistics are important for interpreting the remainder of the experiment. Because the prompts were intentionally authored using the rubric, these agreement values should not be read as external validation on naturally occurring prompts. Rather, they provide evidence of \emph{internal coherence}: prompts designed to instantiate the three classes were, in most cases, perceived by independent annotators in ways broadly consistent with the intended categorization.

\begin{table}[!ht]
\centering
\small
\setlength{\tabcolsep}{6pt}
\begin{tabular}{lccccc}
\toprule
$N_{\text{ann}}$ & $N_{\text{items}}$ & Fleiss' $\kappa$ & Pairwise agr. ($\bar{P}$) & Modal agr. (median) & Modal agr. (min--max) \\
\midrule
51 & 30 & 0.414 & 0.612 & 0.716 & 0.431--1.000 \\
\bottomrule
\end{tabular}
\caption{Inter-annotator agreement for the three-class implicature rubric on the authored prompt set. Fleiss' $\kappa$ gives chance-corrected multi-rater agreement, $\bar{P}$ is average pairwise agreement, and modal agreement indicates the proportion of annotators selecting the most common label for each prompt.}
\label{tab:iaa}
\end{table}

\subsubsection{Model agreement with the human interpretation baseline}

\begin{figure}[!ht]
  \centering
  \includegraphics[width=0.7\textwidth]{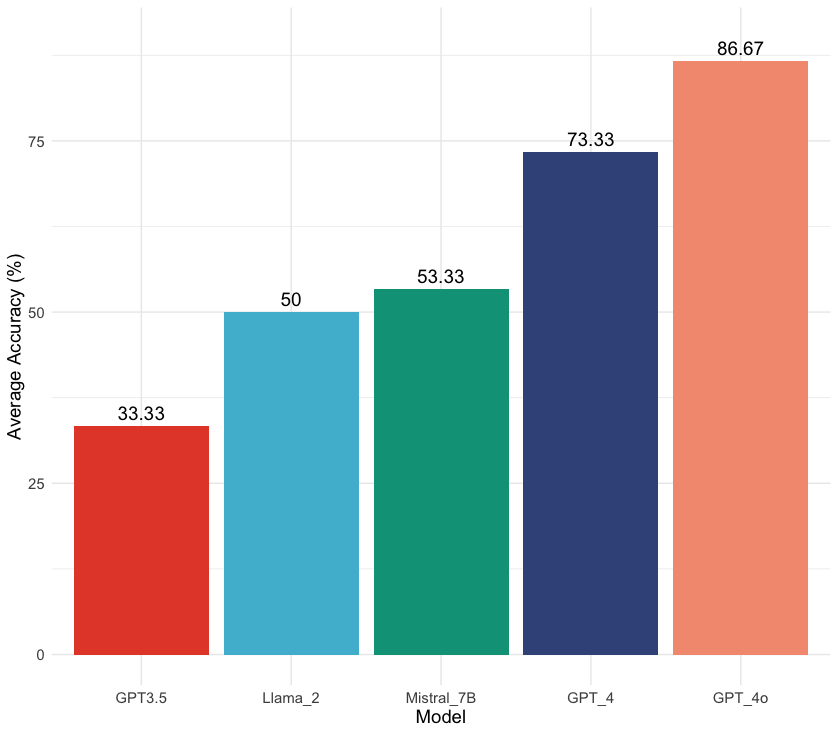}
  \caption{Prompt-level agreement with the modal human label across the five evaluated models.}
  \label{fig:model_accuracy_plot}
\end{figure}

We next benchmarked model interpretations against the modal human label for each prompt. Figure~\ref{fig:model_accuracy_plot} shows prompt-level agreement with the modal human label. Agreement varied substantially across models. GPT-4o achieved the highest agreement at 86.67\% (26/30 prompts), followed by GPT-4 at 73.33\% (22/30), Mistral~7B at 53.33\% (16/30), Llama~2 at 50.00\% (15/30), and GPT-3.5 at 33.33\% (10/30). Overall, larger models aligned more closely with human interpretations, whereas smaller models were more likely to diverge from the modal human judgment.

To complement this prompt-level agreement measure, we also compared model class-score distributions with the empirical human label distributions using category-wise $R^2$ values (Table~\ref{tab:model_r2_values}). This analysis reveals a more nuanced pattern than the headline agreement rates alone. GPT-4 showed the strongest category-level alignment for Information-seeking ($R^2 = 0.89$) and Expressive ($R^2 = 0.97$), while GPT-4o showed the most consistently strong alignment across all three categories ($R^2 = 0.61$ for Information-seeking, $0.69$ for Direction-seeking, and $0.82$ for Expressive). By contrast, smaller models exhibited much weaker alignment overall, particularly in the Direction-seeking class, where both Llama~2 and Mistral~7B yielded very low $R^2$ values ($0.03$). GPT-3.5 showed weak alignment across all categories and was especially poor on Expressive prompts ($R^2 = 0.01$).

At the same time, the category-wise results suggest that smaller models were not uniformly poor. Mistral~7B, for example, showed relatively strong alignment in the Expressive class ($R^2 = 0.83$) despite its modest overall agreement with the human baseline. This indicates that model weaknesses were not evenly distributed across implicature types, and that some classes were more tractable than others even for lower-capacity models.

\begin{table}[!ht]
  \centering
  \small
  \begin{tabular}{@{}llr@{}}
    \toprule
    \textbf{Model} & \textbf{Category} & \textbf{$R^2$} \\
    \midrule
    \multirow{3}{*}{GPT-4o}      & Information-seeking & 0.61 \\
                                 & Direction-seeking   & 0.69 \\
                                 & Expressive          & 0.82 \\
    \multirow{3}{*}{Llama 2}     & Information-seeking & 0.35 \\
                                 & Direction-seeking   & 0.03 \\
                                 & Expressive          & 0.29 \\
    \multirow{3}{*}{GPT-4}       & Information-seeking & 0.89 \\
                                 & Direction-seeking   & 0.71 \\
                                 & Expressive          & 0.97 \\
    \multirow{3}{*}{GPT-3.5}     & Information-seeking & 0.11 \\
                                 & Direction-seeking   & 0.11 \\
                                 & Expressive          & 0.01 \\
    \multirow{3}{*}{Mistral 7B}  & Information-seeking & 0.24 \\
                                 & Direction-seeking   & 0.03 \\
                                 & Expressive          & 0.83 \\
    \bottomrule
  \end{tabular}
  \caption{Category-wise $R^2$ values between model class scores and the empirical human label distributions. Higher values indicate closer alignment with how human annotators distributed labels across the three implicature classes.}
  \label{tab:model_r2_values}
\end{table}

Taken together, Experiment~1 provides two kinds of evidence. First, the three-class rubric was learnable and internally coherent on the authored prompt set, supporting its use as an operational tool in the later experiments. Second, LLMs differed substantially in how closely they aligned with human interpretations of implied intent. The strongest models approached the human baseline more closely, whereas smaller models showed weaker and more uneven alignment, especially for prompts whose primary implied intent involved guidance or affective stance rather than straightforward information need.

\subsection{Experiment~2: Perceived relevance and quality}
\label{sec:exp2_results}

\begin{figure}[!ht]
  \centering
  \includegraphics[width=0.65\textwidth]{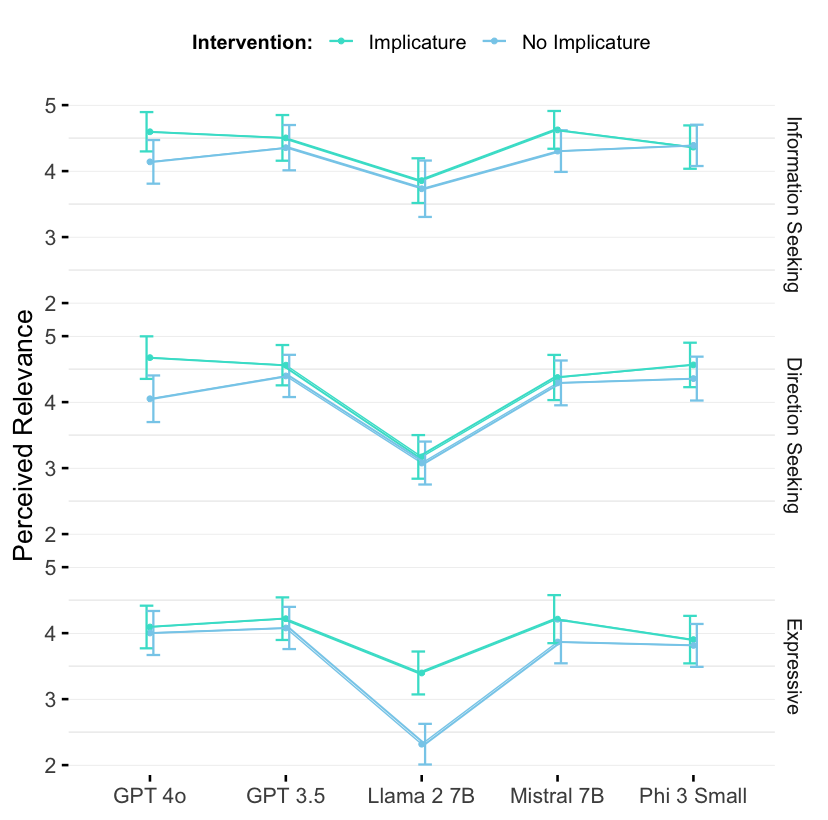}
  \caption{Effects of model, intervention, and class on perceived relevance.}
  \label{fig:relevance_plot}
\end{figure}

\begin{table}[!ht]
  \centering
  \footnotesize
  \begin{tabular}{@{}lrrrrr@{}}
    \toprule
    \textbf{Effect} & \textbf{NumDF} & \textbf{DenDF} & \textbf{F} & \textbf{p} \\
    \midrule
    \textbf{Intervention}            & 1 & 249.99 & 16.64 & $< .001$ \\
    \textbf{Model}                   & 4 & 249.92 & 37.55 & $< .001$ \\
    \textbf{Class}                   & 2 & 247.88 & 21.09 & $< .001$ \\
    Intervention $\times$ Model      & 4 & 250.22 & 1.10  & .357 \\
    Intervention $\times$ Class      & 2 & 253.17 & 0.42  & .658 \\
    Model $\times$ Class             & 8 & 248.14 & 1.70  & .099 \\
    Intervention $\times$ Model $\times$ Class & 8 & 250.69 & 1.63 & .117 \\
    \bottomrule
  \end{tabular}
  \caption{Type~III tests from the linear mixed-effects model for perceived relevance (Satterthwaite's method).}
  \label{tab:anova_relevance}
\end{table}

\begin{figure}[!ht]
  \centering
  \includegraphics[width=0.65\textwidth]{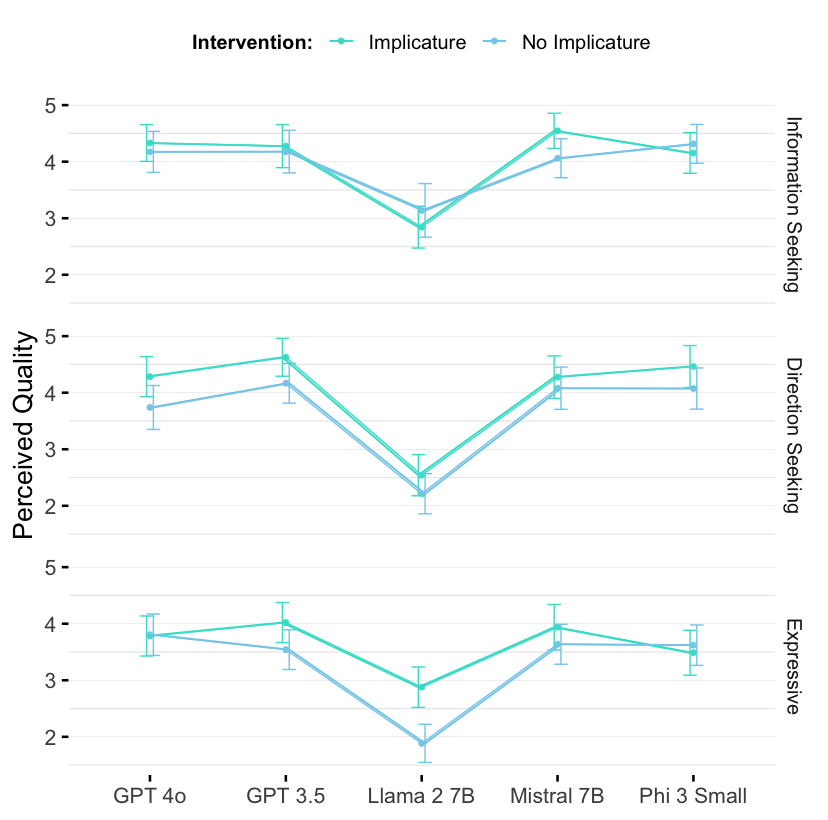}
  \caption{Effects of model, intervention, and class on perceived quality.}
  \label{fig:quality_plot}
\end{figure}

\begin{table}[!ht]
  \centering
  \footnotesize
  \begin{tabular}{@{}lrrrrr@{}}
    \toprule
    \textbf{Effect} & \textbf{NumDF} & \textbf{DenDF} & \textbf{F} & \textbf{p} \\
    \midrule
    \textbf{Intervention}            & 1 & 253.67 & 13.06 & $< .001$ \\
    \textbf{Model}                   & 4 & 252.65 & 63.50 & $< .001$ \\
    \textbf{Class}                   & 2 & 249.59 & 19.90 & $< .001$ \\
    Intervention $\times$ Model      & 4 & 252.61 & 0.68  & .605 \\
    Intervention $\times$ Class      & 2 & 256.31 & 1.84  & .160 \\
    Model $\times$ Class             & 8 & 250.50 & 1.58  & .132 \\
    Intervention $\times$ Model $\times$ Class & 8 & 253.36 & 1.70 & .100 \\
    \bottomrule
  \end{tabular}
  \caption{Type~III tests from the linear mixed-effects model for perceived quality (Satterthwaite's method).}
  \label{tab:anova_quality}
\end{table}

Experiment~2 evaluated whether implicature-aware prompting improved how participants judged LLM responses on two dimensions: perceived relevance and perceived quality. Figures~\ref{fig:relevance_plot} and~\ref{fig:quality_plot} show the model-estimated means across prompting conditions, models, and implicature classes, while Tables~\ref{tab:anova_relevance} and~\ref{tab:anova_quality} report the corresponding Type~III tests from the linear mixed-effects models.

For perceived relevance, there were significant main effects of prompting condition, model, and implicature class. Responses generated under the implicature-aware condition received higher relevance ratings overall than responses generated under the baseline condition, $F(1, 249.99) = 16.64$, $p < .001$. Relevance ratings also differed significantly across models, $F(4, 249.92) = 37.55$, $p < .001$, and across implicature classes, $F(2, 247.88) = 21.09$, $p < .001$. No interaction terms reached conventional significance, including intervention by model, $F(4, 250.22) = 1.10$, $p = .357$; intervention by class, $F(2, 253.17) = 0.42$, $p = .658$; model by class, $F(8, 248.14) = 1.70$, $p = .099$; and the three-way interaction, $F(8, 250.69) = 1.63$, $p = .117$.

A similar pattern emerged for perceived quality. Responses generated under the implicature-aware condition were rated higher in quality overall than baseline responses, $F(1, 253.67) = 13.06$, $p < .001$. Quality ratings also differed significantly across models, $F(4, 252.65) = 63.50$, $p < .001$, and across implicature classes, $F(2, 249.59) = 19.90$, $p < .001$. Again, none of the interaction terms were significant: intervention by model, $F(4, 252.61) = 0.68$, $p = .605$; intervention by class, $F(2, 256.31) = 1.84$, $p = .160$; model by class, $F(8, 250.50) = 1.58$, $p = .132$; and the three-way interaction, $F(8, 253.36) = 1.70$, $p = .100$.

As shown in Figures~\ref{fig:relevance_plot} and~\ref{fig:quality_plot}, implicature-aware prompting generally yielded higher ratings across the evaluated response set. GPT-4o and Mistral~7B tended to receive the highest ratings overall, whereas Llama~2~7B tended to receive the lowest. Because the interaction terms were not significant, the most defensible interpretation is that implicature-aware prompting improved perceived relevance and perceived quality broadly across models and classes, rather than benefiting only a specific subset of conditions.

As descriptive follow-up analyses, we estimated marginal means for prompting condition within each implicature class. For perceived relevance, Holm-adjusted contrasts indicated significant implicature-aware advantages for Direction-seeking items ($\Delta = 0.246$, $p = .047$) and Expressive items ($\Delta = 0.356$, $p = .004$), but not for Information-seeking items ($\Delta = 0.214$, $p = .084$). For perceived quality, the implicature-aware condition was significantly higher for Direction-seeking items ($\Delta = 0.400$, $p = .003$) and Expressive items ($\Delta = 0.329$, $p = .015$), but not for Information-seeking items ($\Delta = 0.069$, $p = .614$). Because the overall intervention-by-class interaction was not significant, these class-wise contrasts should be interpreted as descriptive follow-up results rather than strong evidence of differential intervention effects across classes.

Taken together, Experiment~2 shows that making communicative intent explicit through implicature-aware prompting improved both perceived relevance and perceived quality. Ratings also differed substantially by model and implicature class, but the prompting advantage itself was broadly distributed across the evaluated response set. Full fixed-effect estimates, confidence intervals and variance components for the mixed-effects models are reported in Appendix~A.

\subsection{Experiment~3: Targeted forced-choice follow-up}

\begin{figure}[ht]
    \centering
    \includegraphics[width=0.5\textwidth]{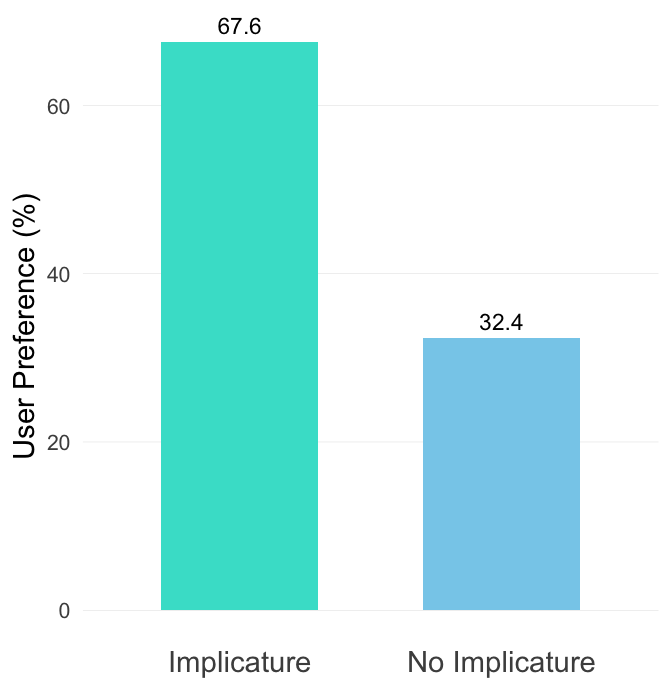}
    \caption{Preference distribution in Experiment~3 across the targeted comparison set.}
    \label{fig:fig2}
\end{figure}

Experiment~3 used a two-alternative forced-choice task with a new participant sample to examine whether selected response contrasts from Experiment~2 also translated into direct user preference when both responses were shown side by side. After deduplicating three duplicate response events in the raw log, the final analysis included 290 valid participant--item choices. Figure~\ref{fig:fig2} summarizes the resulting preference distribution across the targeted comparison set.

Across these valid choices, participants selected the implicature-aware response 67.6\% of the time (196/290), compared with 32.4\% for the baseline response (94/290). An exact binomial test showed that this proportion was significantly above chance (50\%), $p = 1.04 \times 10^{-9}$.
This result should be interpreted in light of the design of Experiment~3. Because the response pairs were selected post hoc from Experiment~2, the forced-choice outcome does not provide an independent estimate of overall user preference across the full prompt set. Rather, it shows that within this targeted comparison set, implicature-aware responses were chosen more often when users directly compared them with baseline alternatives.
Taken together with Experiment~2, these findings suggest that the rating advantages associated with implicature-aware prompting also translated into direct pairwise preference on the selected comparison items.

\section{Discussion}

\subsection{Experiment 1: Implicature Interpretation and Model--Human Alignment}

Experiment~1 benchmarked five large language models (LLMs) against a human interpretation baseline for conversational implicature. The results reveal clear variation in model--human alignment. GPT-4o showed the strongest agreement with the modal human label (86.67\%), followed by GPT-4 (73.33\%), whereas Mistral~7B (53.33\%), Llama~2 (50.00\%), and GPT-3.5 (33.33\%) showed substantially weaker alignment. These findings are broadly consistent with recent work suggesting that larger, more capable models are better able to approximate human pragmatic judgments \cite{bojic2025does, ruis2023goldilocks}.

The category-wise analyses refine this picture. GPT-4 showed particularly strong alignment with the human label distributions for Information-seeking and Expressive prompts, while GPT-4o showed consistently strong alignment across all three implicature classes. In the Expressive category, for example, the model-to-human correspondence was especially high for GPT-4 ($R^2 = 0.97$) and GPT-4o ($R^2 = 0.82$), suggesting that the strongest models were often able to track not only implied informational needs, but also affective and evaluative nuances. At the same time, this pattern should not be taken as evidence of fully human-like pragmatic competence; rather, it indicates closer approximation to the human interpretation baseline on this authored prompt set.

By contrast, the smaller models were less aligned with human interpretations overall, and their weaknesses were not uniform across categories. GPT-3.5 performed poorly across all classes, while Llama~2 showed weaker alignment particularly in the Direction-seeking and Expressive categories. This pattern is consistent with prior work suggesting that smaller or less instruction-tuned models are more likely to default to literal interpretations or rely on rigid heuristics when indirect meaning must be inferred \cite{cho-ismkim99-skku-edu-2024-pragmatic}. At the same time, the results also show that smaller models are not uniformly poor: for example, Mistral~7B showed relatively strong alignment in the Expressive category despite lower overall agreement.

From an HCI perspective, these findings matter because they suggest that pragmatic alignment is not evenly distributed across current model families. In applications where subtle intent recognition is important---such as digital assistants, tutoring systems, or collaborative agents---weaker implicature interpretation may translate into responses that feel overly literal, misaligned, or insufficiently socially attuned. Experiment~1 therefore provides an important motivation for the later user-facing experiments: if models differ substantially in how they interpret implied intent, those differences may carry through into perceived response quality during interaction.

\subsection{Experiment 2: Perceived Relevance and Quality of LLM Responses}

Experiment~2 shifted the focus from implicature interpretation to user-facing response evaluation. The central result is straightforward: making communicative intent explicit through implicature-aware prompting improved both perceived relevance and perceived quality. Participants rated responses generated under the implicature-aware condition more positively overall, and the mixed-effects analyses showed significant main effects of intervention, model, and implicature class on both outcome measures.

At the same time, the corrected models did \emph{not} show significant interaction effects between intervention and model or between intervention and class. This means that the most defensible interpretation is not that the prompting intervention benefited only particular models or only particular classes, but that it improved evaluations broadly across the response set. The descriptive follow-up contrasts suggest that the gains were especially visible for Direction-seeking and Expressive prompts, whereas Information-seeking prompts showed smaller or nonsignificant differences. However, because the overall intervention-by-class interaction was not significant, these class-wise patterns should be interpreted cautiously as descriptive follow-up results rather than strong evidence of differential effects by class.

This finding is important for HCI because it supports a practical design claim. Our results do not show that current models independently infer implicature in a fully human-like way. Rather, they show that when communicative intent is surfaced explicitly through prompting, users judge the resulting responses as better. This suggests that implicature-aware prompting can function as a lightweight intervention for improving interaction quality without requiring retraining or architectural changes to the model. In that sense, the contribution is less about intrinsic model competence and more about how linguistic theory can inform the design of user-facing prompting strategies \cite{ruis2023goldilocks, sravanthi2024pub}.

The differences across models remain important. GPT-4o and Mistral~7B appeared among the highest-rated models overall, whereas Llama~2~7B tended to receive the lowest ratings. Yet because the intervention-by-model interaction was not significant, the data do not support the stronger claim that implicature-aware prompting disproportionately benefits smaller models. Instead, the safer conclusion is that model capability still matters for perceived response quality, but that the prompting intervention improves judgments broadly across the evaluated model set.

For HCI practice, these results point toward a concrete design implication: response quality depends not only on the underlying model, but also on whether the system is guided toward the user's implied communicative goal. Designers of conversational systems should therefore consider prompt strategies that make likely intent explicit, especially in settings where users often communicate indirectly, emotionally, or with incomplete requests. More broadly, the findings reinforce the value of evaluating language-based systems not only in terms of task success or benchmark accuracy, but also in terms of how users perceive the relevance and quality of responses in context.

\subsection{Experiment 3: Targeted forced-choice follow-up}

Experiment~3 used a forced-choice paradigm to examine whether the selected contrasts from Experiment~2 also translated into direct user preference when both response alternatives were shown side by side. Across the targeted comparison set, participants selected the implicature-aware response 67.6\% of the time, compared with 32.4\% for the baseline response. This preference pattern is consistent with the relevance and quality advantages observed in Experiment~2 and suggests that, on the selected response pairs, implicature-aware prompting improved not only scalar evaluations but also direct comparative choice.

At the same time, this result should be interpreted with appropriate caution. Because the response pairs in Experiment~3 were selected post hoc from Experiment~2, the forced-choice outcome does not provide an independent estimate of overall user preference across the full prompt set. Rather, it shows that within this targeted comparison set, implicature-aware responses were chosen more often when users directly compared them against baseline alternatives. The contribution of Experiment~3 is therefore narrower but still important: it demonstrates that the kinds of advantages observed in Experiment~2 can also become visible in direct side-by-side judgments.

The example comparisons are best understood as illustrative cases rather than as direct evidence of participants' reasons for choice (illustrative examples from a preferred pair in the targeted comparison set). In the Expressive example (Fig.~\ref{fig:expressive}), the implicature-aware response appears more affect-sensitive and more directly aligned with the emotional framing of the prompt, whereas the baseline response reads as more detached and descriptive. In the Information-seeking example (Fig.~\ref{fig:information_seeking}), the implicature-aware response appears more focused and structured, while the baseline response provides more diffuse information that may require greater effort to parse. In the Direction-seeking example (Fig.~\ref{fig:direction_seeking}), the implicature-aware response appears more stepwise and action-oriented, whereas the baseline response is comparatively less organized. These examples help illustrate the kinds of response characteristics that may underlie the observed preference pattern, but they should be interpreted as author analyses of the response texts rather than measured participant explanations.

From an HCI perspective, Experiment~3 strengthens the practical significance of the earlier findings. Even when benchmark-style measures or scalar ratings suggest only moderate differences, direct user choice can reveal which response style better aligns with user expectations in context. For interactive systems such as assistants, support agents, or collaborative tools, this matters because users ultimately experience responses comparatively: they care not only whether a response is acceptable, but whether it feels more appropriate, more helpful, or more attuned to the implied intent of the prompt. Experiment~3 therefore complements Experiment~2 by showing that, on selected comparison items, implicature-aware prompting advantages were also reflected in direct user preference.

\subsection{Broader Implications}

Taken together, the three experiments support a broader HCI argument: implicature is a useful and practically consequential test case for understanding how linguistic theory can inform the design and evaluation of human--LLM interaction. At the same time, the findings point to a more nuanced picture than a simple claim that larger models ``understand'' implicature or that prompting alone solves pragmatic alignment.

Experiment~1 shows that model capacity remains strongly associated with alignment to human interpretations of implied intent. GPT-4o and GPT-4 showed substantially stronger agreement with the human interpretation baseline than GPT-3.5, Llama~2, and Mistral~7B, which is consistent with prior work showing that larger models tend to perform better on implicature and broader pragmatics benchmarks \cite{ruis2023goldilocks,cho-ismkim99-skku-edu-2024-pragmatic,anuranjana2024survey}. At the same time, the category-wise analyses complicate simple ``near-human'' claims: strengths were not evenly distributed across implicature classes, and some smaller models showed relative strengths in particular categories despite weaker overall alignment. For HCI, this suggests that model selection matters not only in terms of raw performance, but also in terms of which kinds of implied meaning an application must handle reliably.

Experiments~2 and~3 shift the emphasis from model capability in the abstract to the practical value of design interventions. Our results show that making communicative intent explicit through implicature-aware prompting improves perceived relevance and perceived quality, and that selected comparison items from Experiment~2 also translate into direct pairwise preference in Experiment~3. This extends prior work on instruction tuning, prompt formulation, and pragmatic evaluation \cite{ruis2023goldilocks,webson2022prompt,sravanthi2024pub} by showing that a lightweight prompting intervention can produce measurable user-facing benefits without retraining the underlying model. Importantly, the final mixed-effects analyses in Experiment~2 did not show reliable intervention-by-model or intervention-by-class interactions. This means the safest conclusion is not that the intervention disproportionately benefits only smaller models or only certain implicature classes, but that it improves judgments broadly across the evaluated response set. For HCI, that is still a valuable result: prompt design can function as a practical lever for improving interaction quality even when deeper model limitations remain.

Experiment~3 highlights the value of including direct user preference alongside scalar rating measures. Benchmark-style metrics and even Likert ratings can identify response differences, but forced-choice judgments reveal whether those differences matter when users directly compare alternatives. In our targeted follow-up, implicature-aware responses were preferred more often than baseline responses, suggesting that the gains observed in Experiment~2 were not only statistically detectable but also visible at the level of direct comparison. That said, the interpretation of Experiment~3 must remain appropriately narrow. Because the response pairs were selected post hoc from Experiment~2, the forced-choice result should not be read as a population-level estimate of overall preference across the full prompt set. Instead, it provides targeted evidence that the prompting advantages identified in the rating study also translated into direct user choice on selected items. Even with that caveat, the result reinforces an important HCI point: users care not only whether a response is acceptable, but whether it feels more appropriate, more helpful, and more attuned to their implied communicative intent.

Our results also point to a gap between successful response behavior and deeper pragmatic reasoning. Experiment~1 shows that some models align more closely with human interpretations when explicitly tasked with classifying implicature, while Experiments~2 and~3 show that prompting can improve the perceived quality of generated responses. Yet neither finding should be taken as evidence that models possess stable, human-like pragmatic reasoning. Rather, the results are more consistent with a weaker claim: current models can be guided toward better pragmatic behavior when communicative intent is made explicit, but this does not imply that they reliably infer such intent unaided in open interaction. Current results also suggest that scale alone is not the whole story. Although larger models showed stronger alignment with the human interpretation baseline in Experiment~1, the user-facing gains in Experiments~2 and~3 came from making communicative intent explicit through prompting rather than from scale alone. This suggests that simply scaling transformer-based models, while often beneficial \cite{kaplan2020scalinglawsneurallanguage}, may be insufficient for achieving robust, human-like implicature handling without additional pragmatic scaffolding or new forms of reasoning support. This interpretation is consistent with broader critiques arguing that LLM success on reasoning-like tasks may often reflect statistical pattern matching rather than robust underlying reasoning \cite{krakauer2025large, stechly2025beyond, kambhampati2025stop, west2023generative, shojaee2025illusion}. It also aligns with work showing that models can arrive at plausible answers while still offering shallow, unstable, or post hoc explanations \cite{hota2025conscience,hota2025nomiclaw,webson2022prompt,srivastava2023beyond,lanham2023measuring}. For HCI, the implication is that prompt-level gains are useful, but they should not be conflated with genuine pragmatic competence.

More broadly, the study supports the view that linguistic phenomena such as implicature, politeness, repair, stance, and figurative language should not be treated as peripheral concerns in human--AI interaction. As language becomes an increasingly central interface, HCI must engage more directly with pragmatics, discourse analysis, and related linguistic traditions in order to design, evaluate, and theorize interactive systems. Our contribution is modest in scope but concrete in implication: implicature is not only a benchmark problem for NLP, but also a user-experience issue for HCI. We therefore see this work as part of a broader agenda for bringing linguistic theory into HCI in a more operational and evaluative way. The key lesson is not simply that LLMs can sometimes approximate implicature, but that the handling of implied meaning measurably shapes how users perceive and prefer system responses. That, in turn, makes linguistic alignment a central concern for the future of language-based interaction design.

\subsection{Discussion Summary}

Taken together, our findings contribute to both HCI and the growing literature on pragmatics in LLMs. The results support a clear but modest conclusion: implicature matters for human--LLM interaction, and making communicative intent explicit through prompting can measurably improve how users perceive and choose model responses. At the same time, the gap between prompt-guided performance and genuine human-like pragmatic competence remains substantial.

More specifically, the paper makes five main contributions. It introduces an HCI-oriented operational taxonomy of three implicature classes, grounded in Grice's cooperative principle and Searle's speech act theory, and shows that this distinction can be applied with reasonable consistency on an authored prompt set. It then benchmarks model--human alignment in implicature interpretation, showing that larger models more closely approximate the human interpretation baseline, whereas smaller models are less consistent and more uneven across classes. The paper further demonstrates that implicature-aware prompting improves perceived relevance and perceived quality in user-facing response evaluation, extending prior benchmark-focused work \cite{anuranjana2024survey,ruis2023goldilocks} toward HCI-relevant experience measures. In a targeted forced-choice follow-up, selected comparison items from the rating study also translate into direct user preference, suggesting that the benefits of implicature-aware prompting remain visible when users compare alternatives side by side. Finally, the study highlights an important conceptual distinction: current models can often be guided toward better pragmatic behavior when communicative intent is made explicit, but such gains should not be mistaken for stable, unaided pragmatic reasoning.

These findings also carry concrete design implications for HCI. They suggest that pragmatic prompting can serve as a lightweight intervention for improving interaction quality, that model choice still matters for how indirect meaning is handled, and that user-centered evaluation should look beyond benchmark accuracy to include perceived relevance, quality, and direct preference. More broadly, the work argues for treating linguistic alignment as a central concern for the future of language-based interaction design.

\section{Future Work and Conclusion}

\subsection{Limitations and Future Work}

The findings should be interpreted in light of several limitations, each of which points to an important direction for future research. First, the intervention we evaluate is \emph{implicature-aware prompting}, not unaided implicature inference. The improvements observed in Experiments~2 and~3 therefore reflect the value of making likely communicative intent explicit to the model through brief class-based guidance. They should not be taken as evidence that current models independently and consistently perform human-like pragmatic reasoning. A natural next step is therefore to move beyond prompt-level scaffolding and test end-to-end architectures in which implicature is inferred automatically and then used to guide response generation. For example, a separate component could predict a likely implicature class, or a probabilistic distribution over classes, and pass that information to the response model as lightweight pragmatic guidance.

A related limitation is that the present setup functions as a Wizard-of-Oz-style approximation of an implicature-aware interaction pipeline. In our experiments, the relevant implicature class was supplied through brief class-definition guidance rather than detected automatically at runtime. This was appropriate for isolating whether surfacing communicative intent can improve user evaluations, but it does not capture the uncertainty and error propagation that would arise in a deployable system. Future work should therefore evaluate complete pipelines in which implicature detection, uncertainty handling, and response generation are integrated, and should examine when clarification is preferable to confident but potentially incorrect pragmatic inference.

The scope of the taxonomy also warrants caution. The three-class scheme used here was intentionally designed as a lightweight, HCI-oriented operationalization rather than a comprehensive linguistic account of implicature. Moreover, the experimental prompts were authored using the rubric itself. Experiment~1 therefore demonstrates learnability and internal coherence on this authored stimulus set, but not external validity across naturally occurring prompts or the broader range of pragmatic phenomena found in everyday interaction. Future work should test the taxonomy on more naturalistic datasets and examine whether broader or more fine-grained pragmatic schemes yield stronger explanatory or design value.

The preference results from Experiment~3 should also be interpreted narrowly. Those items were sampled post hoc from a high-contrast subset of the Experiment~2 response pool rather than independently from the full prompt set. As a result, the preference estimate should be understood as evidence for the selected comparison set rather than as an unbiased estimate of overall user preference across all prompts. A useful next step would be to run larger confirmatory preference studies with prospectively sampled items and preregistered selection procedures.

Further limitations concern participant and language scope. Our participants were English-speaking adults recruited online, and all prompts and responses were in English. Because implicature depends heavily on linguistic convention, discourse norms, and social context, the findings should not be assumed to generalize directly to multilingual or cross-cultural settings. Broader evaluation across languages, cultures, and discourse communities is therefore essential for assessing whether the taxonomy and the observed prompting benefits transfer beyond the present context.

The study is also limited in pragmatic breadth. We focused on three implicature classes---information-seeking, direction-seeking, and expressive---because they are common and tractable in human--LLM interaction. This leaves out other important forms of pragmatic meaning, including politeness strategies, sarcasm, irony, figurative language, stance, and conversational repair. Extending the analysis to these phenomena would help determine whether the benefits of explicit pragmatic scaffolding generalize across a wider range of interactional situations.

Another limitation is ecological validity. The experiments were short, online, and task-based, and therefore do not capture the full longitudinal, relational, and multimodal character of real-world interaction. In practice, pragmatic interpretation unfolds across longer conversations and often depends on shared history, trust, adaptation, prosody, gesture, layout, and other contextual cues. Future work should therefore examine implicature-sensitive behavior in longitudinal and field-based settings, as well as in multimodal interaction contexts where meaning is distributed across more than text alone.

Finally, we did not directly examine the ethical risks of inferring implied intent. Systems that attempt to read between the lines may over-interpret ambiguous utterances, infer more than users intend to disclose, or produce responses that feel presumptuous, manipulative, or privacy-invasive. If pragmatic sensitivity is to become a trustworthy feature of conversational systems, future research must address transparency, user control, contestability, and uncertainty communication alongside performance gains.

\subsection{Conclusion}

This paper examined how implicature can be operationalized and evaluated in human--LLM interaction. Across three experiments, we found that larger models aligned more closely with a human interpretation baseline for implicature, whereas smaller models showed weaker and less consistent alignment across classes. We also found that making communicative intent explicit through implicature-aware prompting improved perceived relevance and perceived quality in user-facing evaluation. In a targeted forced-choice follow-up, these advantages also translated into direct user preference for implicature-aware responses on the selected comparison set.

These findings carry a practical implication for HCI and conversational system design. They suggest that the handling of implied meaning is not merely a theoretical issue in pragmatics or a benchmark problem in NLP, but a factor that measurably shapes user experience. A lightweight prompting intervention based on pragmatic theory was sufficient to improve how responses were perceived, even without retraining the underlying models. This points to a useful design lesson: systems should not treat all user inputs as literal information requests, but should instead be designed to account for the possibility that users are seeking guidance, expressing affect, or implying needs indirectly. Such sensitivity may be especially valuable in assistants, support systems, educational tools, and other interactive settings where appropriateness, responsiveness, and tone matter as much as factual correctness.

At the same time, these results should not be read as evidence that current models possess stable, human-like pragmatic competence. The observed gains came from pragmatic scaffolding rather than reliable unaided implicature inference in open-ended interaction.
The contribution of this work is therefore not to claim that implicature has been solved, but to show that explicitly representing likely communicative intent can improve user-facing outcomes and can provide a productive bridge between linguistic theory and interaction design.

Hence, the study supports a modest but important conclusion: implicature matters for HCI. Systems that better handle implied meaning can produce responses that users experience as more relevant, more appropriate, and more aligned with the communicative situation. Advancing this area will require broader operationalizations, more realistic evaluations, and more careful attention to the ethical and design consequences of inferring beyond literal content, but the present findings suggest that this is a promising direction for the design of language-based interactive systems.

\section*{Declaration on the use of Generative AI}

The author(s) acknowledge the use of GenAI tools (specifically, OpenAI’s ChatGPT 4x and 5x series) in the preparation of this manuscript. These tools were employed solely for formatting assistance, language polishing, and other editorial tasks (e.g., improving clarity, correcting grammar, and ensuring consistent style). All substantive ideas, analyses, conceptual contributions, and interpretations presented in this paper are the original work of the authors, who bear full responsibility for its content.
After using these tool(s)/service(s), the author(s) reviewed and edited the content as needed and take(s) full responsibility for the publication’s content.

\section*{Data and Code Availability}

Upon acceptance, we will release the authored prompt set, model-generated response materials, experimental instructions, prompting templates, anonymized derived datasets, and analysis code used to reproduce the reported results and figures. We will also document model identifiers, prompting conditions, and the final analysis scripts used for each experiment.

\section*{Author Contributions}

Asutosh Hota contributed to the conceptualization, methodology, investigation, formal analysis, data curation, visualization, and writing of the manuscript, including the original draft and subsequent review and editing. Jussi P.\ P.\ Jokinen contributed to the conceptualization, methodology, supervision, validation, and review and minor editing of the manuscript. Both authors contributed to the interpretation of the results, critically revised the manuscript, approved the final version to be sent for review and agree to be accountable for all aspects of the work.

\bibliographystyle{plain}
\bibliography{refs} 

@article{chierchia2004scalar,
  title={Scalar implicatures, polarity phenomena, and the syntax/pragmatics interface},
  author={Chierchia, Gennaro and others},
  journal={Structures and beyond},
  volume={3},
  pages={39--103},
  year={2004},
  publisher={Oxford, UK}
}

@article{sauerland2012computation,
  title={The computation of scalar implicatures: Pragmatic, lexical or grammatical?},
  author={Sauerland, Uli},
  journal={Language and Linguistics Compass},
  volume={6},
  number={1},
  pages={36--49},
  year={2012},
  publisher={Wiley Online Library}
}

@article{recanati1989pragmatics,
  title={The pragmatics of what is said},
  author={Recanati, Fran{\c{c}}ois},
  year={1989}
}

@article{chierchia2012grammatical,
  title={The grammatical view of scalar implicatures and the relationship between semantics and pragmatics},
  author={Chierchia, Gennaro and Fox, Danny and Spector, Benjamin},
  journal={Semantics: An international handbook of natural language meaning},
  volume={3},
  pages={2297--2332},
  year={2012}
}

@incollection{spector2007aspects,
  title={Aspects of the pragmatics of plural morphology: On higher-order implicatures},
  author={Spector, Benjamin},
  booktitle={Presupposition and implicature in compositional semantics},
  pages={243--281},
  year={2007},
  publisher={Springer}
}

@article{recanati2003embedded,
  title={Embedded implicatures},
  author={Recanati, Fran{\c{c}}ois},
  journal={Philosophical perspectives},
  volume={17},
  pages={299--332},
  year={2003},
  publisher={JSTOR}
}

@book{grice1991studies,
  title={Studies in the Way of Words},
  author={Grice, Paul},
  year={1991},
  publisher={Harvard University Press}
}

@incollection{grice1975logic,
  title={Logic and conversation},
  author={Grice, Herbert P},
  booktitle={Speech acts},
  pages={41--58},
  year={1975},
  publisher={Brill}
}

@book{huang2017oxford,
  title={The Oxford handbook of pragmatics},
  author={Huang, Yan},
  year={2017},
  publisher={Oxford University Press}
}

@article{ruis2023goldilocks,
  title={The goldilocks of pragmatic understanding: Fine-tuning strategy matters for implicature resolution by llms},
  author={Ruis, Laura and Khan, Akbir and Biderman, Stella and Hooker, Sara and Rockt{\"a}schel, Tim and Grefenstette, Edward},
  journal={Advances in Neural Information Processing Systems},
  volume={36},
  pages={20827--20905},
  year={2023}
}

@incollection{searle1975indirect,
  title={Indirect speech acts},
  author={Searle, John R},
  booktitle={Speech acts},
  pages={59--82},
  year={1975},
  publisher={Brill}
}

@article{bojic2025does,
  title={Does GPT-4 surpass human performance in linguistic pragmatics?},
  author={Boji{\'c}, Ljubi{\v{s}}a and Kova{\v{c}}evi{\'c}, Predrag and {\v{C}}abarkapa, Milan},
  journal={Humanities and Social Sciences Communications},
  volume={12},
  number={1},
  pages={1--10},
  year={2025},
  publisher={Palgrave}
}

@inproceedings{cho-ismkim99-skku-edu-2024-pragmatic,
    title = "Pragmatic inference of scalar implicature by {LLM}s",
    author = "Cho, Ye-eun  and
      Kim, Seong mook",
    editor = "Fu, Xiyan  and
      Fleisig, Eve",
    booktitle = "Proceedings of the 62nd Annual Meeting of the Association for Computational Linguistics (Volume 4: Student Research Workshop)",
    month = aug,
    year = "2024",
    address = "Bangkok, Thailand",
    publisher = "Association for Computational Linguistics",
    url = "https://aclanthology.org/2024.acl-srw.2/",
    doi = "10.18653/v1/2024.acl-srw.2",
    pages = "10--20",
    ISBN = "979-8-89176-097-4"
}

@misc{liu2023trainingsociallyalignedlanguage,
      title={Training Socially Aligned Language Models on Simulated Social Interactions}, 
      author={Ruibo Liu and Ruixin Yang and Chenyan Jia and Ge Zhang and Denny Zhou and Andrew M. Dai and Diyi Yang and Soroush Vosoughi},
      year={2023},
      eprint={2305.16960},
      archivePrefix={arXiv},
      primaryClass={cs.CL},
      url={https://arxiv.org/abs/2305.16960}, 
}

@article{george2020conversational,
  title={Conversational implicatures in English dialogue: Annotated dataset},
  author={George, Elizabeth Jasmi and Mamidi, Radhika},
  journal={Procedia Computer Science},
  volume={171},
  pages={2316--2323},
  year={2020},
  publisher={Elsevier}
}

@inproceedings{anuranjana2024survey,
  title={Survey on Computational Approaches to Implicature},
  author={Anuranjana, Kaveri and Mallepally, Srihitha and Mareddy, Sriharshitha and Shukla, Amit and Mamidi, Radhika},
  booktitle={Proceedings of the 21st International Conference on Natural Language Processing (ICON)},
  pages={224--229},
  year={2024}
}

@article{sravanthi2024pub,
  title={Pub: A pragmatics understanding benchmark for assessing llms' pragmatics capabilities},
  author={Sravanthi, Settaluri Lakshmi and Doshi, Meet and Kalyan, Tankala Pavan and Murthy, Rudra and Bhattacharyya, Pushpak and Dabre, Raj},
  journal={arXiv preprint arXiv:2401.07078},
  year={2024}
}

@inproceedings{yue2024large,
  title={Do large language models understand conversational implicature--a case study with a Chinese sitcom},
  author={Yue, Shisen and Song, Siyuan and Cheng, Xinyuan and Hu, Hai},
  booktitle={China National Conference on Chinese Computational Linguistics},
  pages={402--418},
  year={2024},
  organization={Springer}
}

@inproceedings{liang2019implicit,
  title={Implicit communication of actionable information in human-ai teams},
  author={Liang, Claire and Proft, Julia and Andersen, Erik and Knepper, Ross A},
  booktitle={Proceedings of the 2019 CHI conference on human factors in computing systems},
  pages={1--13},
  year={2019}
}

@inproceedings{serim2019explicating,
  title={Explicating" Implicit Interaction" An examination of the concept and challenges for research},
  author={Serim, Bar{\i}{\c{s}} and Jacucci, Giulio},
  booktitle={Proceedings of the 2019 chi conference on human factors in computing systems},
  pages={1--16},
  year={2019}
}

@inproceedings{nishihata2023human,
  title={Human-like “agents” or “tools”?: Exploring the implicature-of-quantity in HAI},
  author={Nishihata, Chisato and Kobayashi, Harumi and Yasuda, Tetsuya},
  booktitle={Proceedings of the 11th International Conference on Human-Agent Interaction},
  pages={387--389},
  year={2023}
}

@inproceedings{iida2024integrating,
  title={Integrating Large Language Model and Mental Model of Others: Studies on Dialogue Communication Based on Implicature},
  author={Iida, Ayu and Okuoka, Kohei and Fukuda, Satoko and Omori, Takashi and Nakashima, Ryoichi and Osawa, Masahiko},
  booktitle={Proceedings of the 12th International Conference on Human-Agent Interaction},
  pages={260--269},
  year={2024}
}

@inproceedings{kehler2000cognitive,
  title={Cognitive status and form of reference in multimodal human-computer interaction},
  author={Kehler, Andrew},
  booktitle={AAAI/IAAI},
  pages={685--690},
  year={2000}
}

@article{oberlander1995grice,
  title={Grice for graphics: pragmatic implicature in network diagrams},
  author={Oberlander, Jon},
  journal={Information design journal},
  volume={8},
  number={2},
  pages={163--179},
  year={1995},
  publisher={John Benjamins}
}

@article{cong2024manner,
  title={Manner implicatures in large language models},
  author={Cong, Yan},
  journal={Scientific Reports},
  volume={14},
  number={1},
  pages={29113},
  year={2024},
  publisher={Nature Publishing Group UK London}
}

@article{shin2021effects,
  title={The effects of explainability and causability on perception, trust, and acceptance: Implications for explainable AI},
  author={Shin, Donghee},
  journal={International journal of human-computer studies},
  volume={146},
  pages={102551},
  year={2021},
  publisher={Elsevier}
}

@article{saygin2002pragmatics,
  title={Pragmatics in human-computer conversations},
  author={Saygin, Ayse Pinar and Cicekli, Ilyas},
  journal={Journal of Pragmatics},
  volume={34},
  number={3},
  pages={227--258},
  year={2002},
  publisher={Elsevier}
}

@incollection{chierchia2012scalar,
  title={Scalar implicature as a grammatical phenomenon},
  author={Chierchia, Gennaro and Fox, Danny and Spector, Benjamin},
  booktitle={Handb{\"u}cher zur Sprach-und Kommunikationswissenschaft/Handbooks of Linguistics and Communication Science Semantics Volume 3},
  year={2012},
  publisher={de Gruyter}
}

@article{krakauer2025large,
  title={Large Language Models and Emergence: A Complex Systems Perspective},
  author={Krakauer, David C and Krakauer, John W and Mitchell, Melanie},
  journal={arXiv preprint arXiv:2506.11135},
  year={2025}
}

@article{stechly2025beyond,
  title={Beyond semantics: The unreasonable effectiveness of reasonless intermediate tokens},
  author={Stechly, Kaya and Valmeekam, Karthik and Gundawar, Atharva and Palod, Vardhan and Kambhampati, Subbarao},
  journal={arXiv preprint arXiv:2505.13775},
  year={2025}
}

@article{kambhampati2025stop,
  title={Stop Anthropomorphizing Intermediate Tokens as Reasoning/Thinking Traces!},
  author={Kambhampati, Subbarao and Stechly, Kaya and Valmeekam, Karthik and Saldyt, Lucas and Bhambri, Siddhant and Palod, Vardhan and Gundawar, Atharva and Samineni, Soumya Rani and Kalwar, Durgesh and Biswas, Upasana},
  journal={arXiv preprint arXiv:2504.09762},
  year={2025}
}

@article{west2023generative,
  title={The Generative AI paradox:" What it can create, it may not understand"},
  author={West, Peter and Lu, Ximing and Dziri, Nouha and Brahman, Faeze and Li, Linjie and Hwang, Jena D and Jiang, Liwei and Fisher, Jillian and Ravichander, Abhilasha and Chandu, Khyathi and others},
  journal={arXiv preprint arXiv:2311.00059},
  year={2023}
}

@article{shojaee2025illusion,
  title={The illusion of thinking: Understanding the strengths and limitations of reasoning models via the lens of problem complexity},
  author={Shojaee, Parshin and Mirzadeh, Iman and Alizadeh, Keivan and Horton, Maxwell and Bengio, Samy and Farajtabar, Mehrdad},
  journal={arXiv preprint arXiv:2506.06941},
  year={2025}
}

@article{hota2025conscience,
  title={Conscience conflict? Evaluating language models’ moral understanding},
  author={Hota, Asutosh and Jokinen, Jussi PP},
  year={2025}
}

@article{hota2025nomiclaw,
  title={NomicLaw: Emergent Trust and Strategic Argumentation in LLMs During Collaborative Law-Making},
  author={Hota, Asutosh and Jokinen, Jussi PP},
  journal={arXiv preprint arXiv:2508.05344},
  year={2025}
}

@inproceedings{webson2022prompt,
  title={Do prompt-based models really understand the meaning of their prompts?},
  author={Webson, Albert and Pavlick, Ellie},
  booktitle={Proceedings of the 2022 conference of the north american chapter of the association for computational linguistics: Human language technologies},
  pages={2300--2344},
  year={2022}
}

@article{srivastava2023beyond,
  title={Beyond the imitation game: Quantifying and extrapolating the capabilities of language models},
  author={Srivastava, Aarohi and Rastogi, Abhinav and Rao, Abhishek and Shoeb, Abu Awal and Abid, Abubakar and Fisch, Adam and Brown, Adam R and Santoro, Adam and Gupta, Aditya and Garriga-Alonso, Adri and others},
  journal={Transactions on machine learning research},
  year={2023}
}

@article{lanham2023measuring,
  title={Measuring faithfulness in chain-of-thought reasoning},
  author={Lanham, Tamera and Chen, Anna and Radhakrishnan, Ansh and Steiner, Benoit and Denison, Carson and Hernandez, Danny and Li, Dustin and Durmus, Esin and Hubinger, Evan and Kernion, Jackson and others},
  journal={arXiv preprint arXiv:2307.13702},
  year={2023}
}

@misc{kaplan2020scalinglawsneurallanguage,
      title={Scaling Laws for Neural Language Models}, 
      author={Jared Kaplan and Sam McCandlish and Tom Henighan and Tom B. Brown and Benjamin Chess and Rewon Child and Scott Gray and Alec Radford and Jeffrey Wu and Dario Amodei},
      year={2020},
      eprint={2001.08361},
      archivePrefix={arXiv},
      primaryClass={cs.LG},
      url={https://arxiv.org/abs/2001.08361}, 
}

@inproceedings{pang2025understanding,
  title={Understanding the LLM-ification of CHI: Unpacking the Impact of LLMs at CHI through a Systematic Literature Review},
  author={Pang, Rock Yuren and Schroeder, Hope and Smith, Kynnedy Simone and Barocas, Solon and Xiao, Ziang and Tseng, Emily and Bragg, Danielle},
  booktitle={Proceedings of the 2025 CHI Conference on Human Factors in Computing Systems},
  pages={1--20},
  year={2025}
}

@article{shneiderman1982future,
  title={The future of interactive systems and the emergence of direct manipulation},
  author={Shneiderman, Ben},
  journal={Behaviour \& Information Technology},
  volume={1},
  number={3},
  pages={237--256},
  year={1982},
  publisher={Taylor \& Francis}
}

@inproceedings{sun2024generative,
  title={Generative AI in the wild: Prospects, challenges, and strategies},
  author={Sun, Yuan and Jang, Eunchae and Ma, Fenglong and Wang, Ting},
  booktitle={Proceedings of the 2024 CHI Conference on Human Factors in Computing Systems},
  pages={1--16},
  year={2024}
}

@String{Computing = "Computing" }

@String{Computer = "{IEEE} Computer" }

@String{Springer = "Springer-Verlag" }

\pagebreak

\appendix

\section{Appendix}\label{appendix}
\begin{lstlisting}[caption={System prompt for LLM classification of implicature in Experiment 1. This prompt frames the interaction, defines the concept of implication, and introduces the three implicature classes.}, label={lst:exp1-system-prompt}]
Imagine a scenario where 2 people, Person A and B, are texting each other over phone. Person A (Speaker) asks a question to Person B (reader). For this task, you have to assume that you are Person B. You have to read the text and select from the options below what is a suitable implication of the text.

What is an implication? 

An implication is the hidden meaning of the text and might differ from the literal meaning of the sentence. It refers to things that a reader can work out from the way something was said rather than what was actually said.

People process these implications all of the time and are mostly unaware of it.

For example:

Person A: How are you doing? Long time, no see.

Underlying implication: The person here is trying to start a conversation with another person and the underlying implication of this sentence is to get the conversation going.

We have identified 3 types of implications that Person A could have while texting with Person B.

Types of implication:

1. Information Seeking: Asking for information, facts, or knowledge from others. The primary goal is to obtain necessary data or insights. For example, "What is the weather report for the next week?"

2. Direction Seeking: Asking for instructions or directions to perform a specific task or action. It often involves commands, instructions, or requests, leading to an action. For instance, seeking instructions to complete an assignment.

3. Expressive: Communicating feelings, emotions, opinions, or attitudes. The focus is on sharing one's personal state rather than expecting information or action. For example, saying "I'm really happy about the results" expresses one's feelings.

Your task is to read the message as Person B and select the implication class (Information Seeking, Direction Seeking, or Expressive) that best describes the underlying meaning of the prompt provided to you.

\end{lstlisting}

\begin{lstlisting}[caption={System prompt and implicature class conditions used to guide LLM responses in experiment 2. The system prompt establishes the conversational context, while each condition specifies the pragmatic goal for the model's reply.}, label={lst:system-prompt}]
system_prompt = (
    "You are a helpful and context-aware AI assistant. "
    "For each user input, consider the communicative goal specified below "
    "(information seeking, direction seeking, or expressive). "
    "Interpret the user's intent accordingly and generate a response that is sensitive to this context."
)

conditions = {
    "Information Seeking": system_prompt + " The user is seeking information, facts, or knowledge. Your goal is to provide relevant data or insights in response to their query.",
    "Direction Seeking": system_prompt + " The user is seeking instructions or guidance to accomplish a specific task. Your response should provide actionable steps or directions.",
    "Expressive": system_prompt + " The user is expressing emotions, feelings, or attitudes. Your response should acknowledge and engage with the user's personal state or sentiment."
}
\end{lstlisting}

\begin{figure}[ht]
    \centering
    \includegraphics[width=\linewidth]{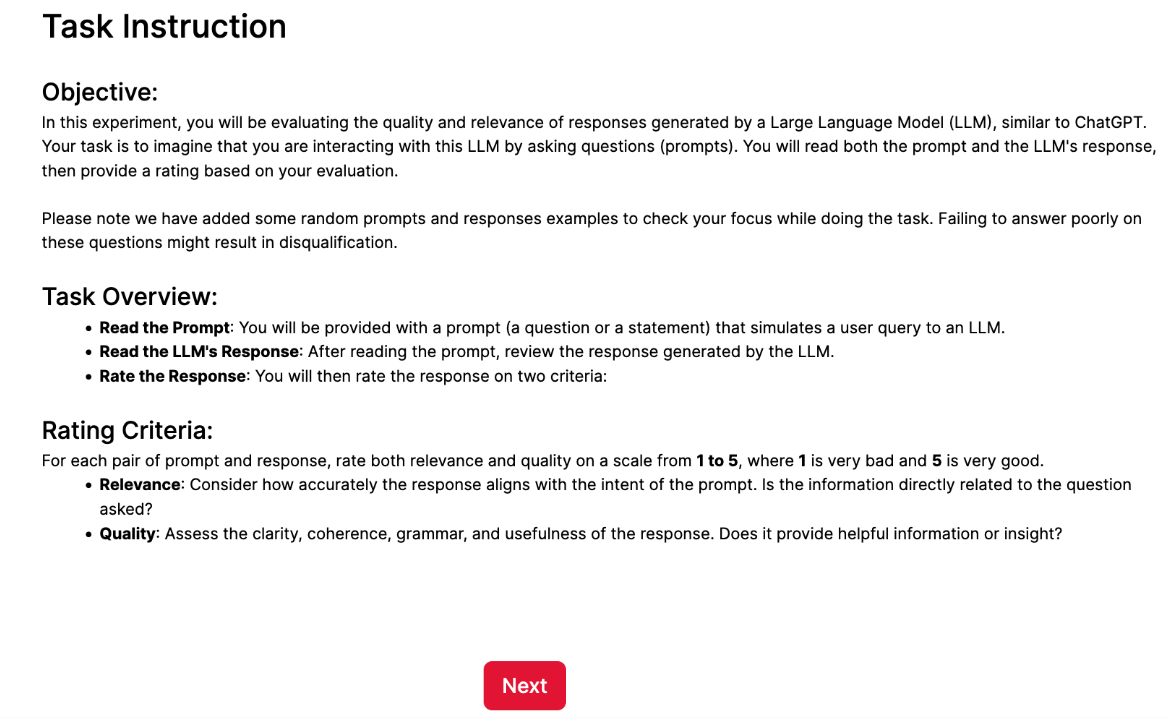}
    \caption{Instructions provided to participants for Experiment 2 (perceived relevance and quality task). Participants were asked to read a prompt and an LLM-generated response, and then rate the response on both relevance and quality using a 5-point Likert scale.}
    \label{fig:exp2-instruction}
\end{figure}

\begin{figure}[t]
    \centering
    \includegraphics[width=\linewidth]{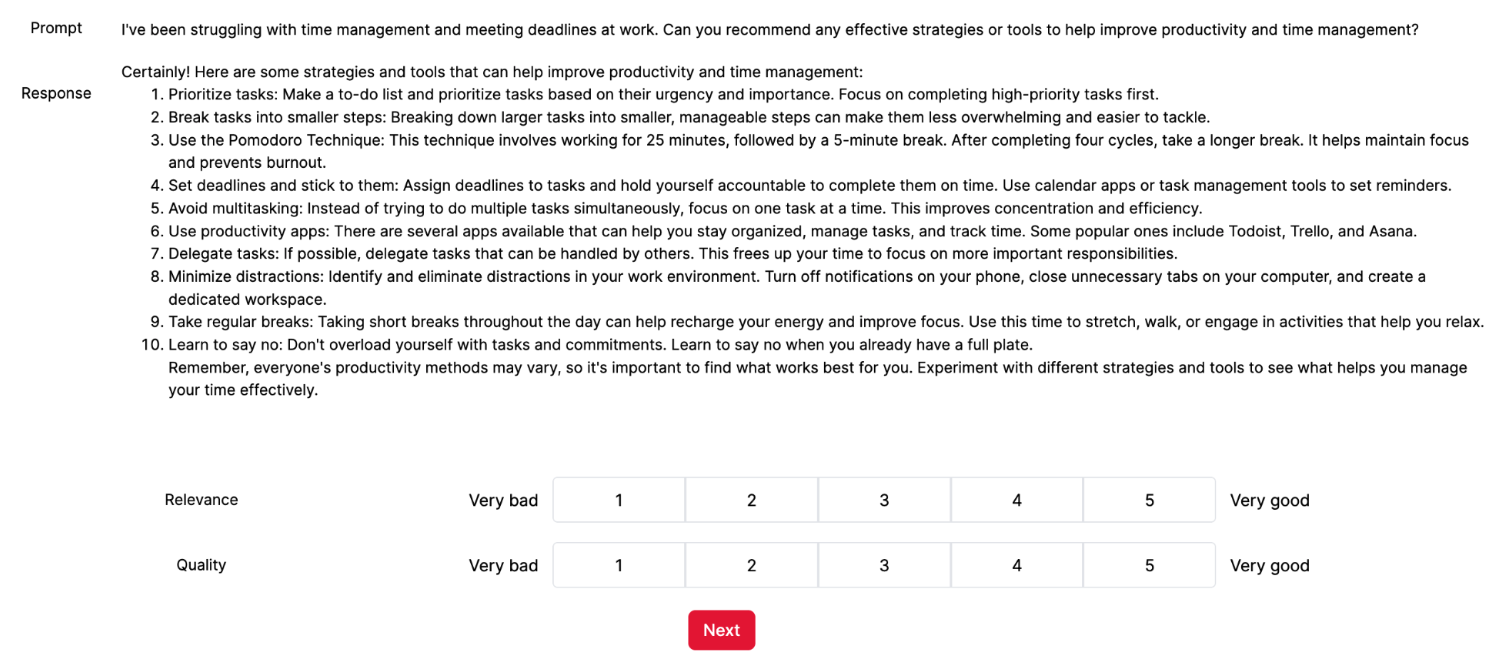}
    \caption{Example prompt and response pair from Experiment 2. Participants rated the relevance (alignment with user intent) and quality (clarity, coherence, usefulness) of the LLM response.}
    \label{fig:exp2-example}
\end{figure}

\begin{figure}[t]
    \centering
    \includegraphics[width=\linewidth]{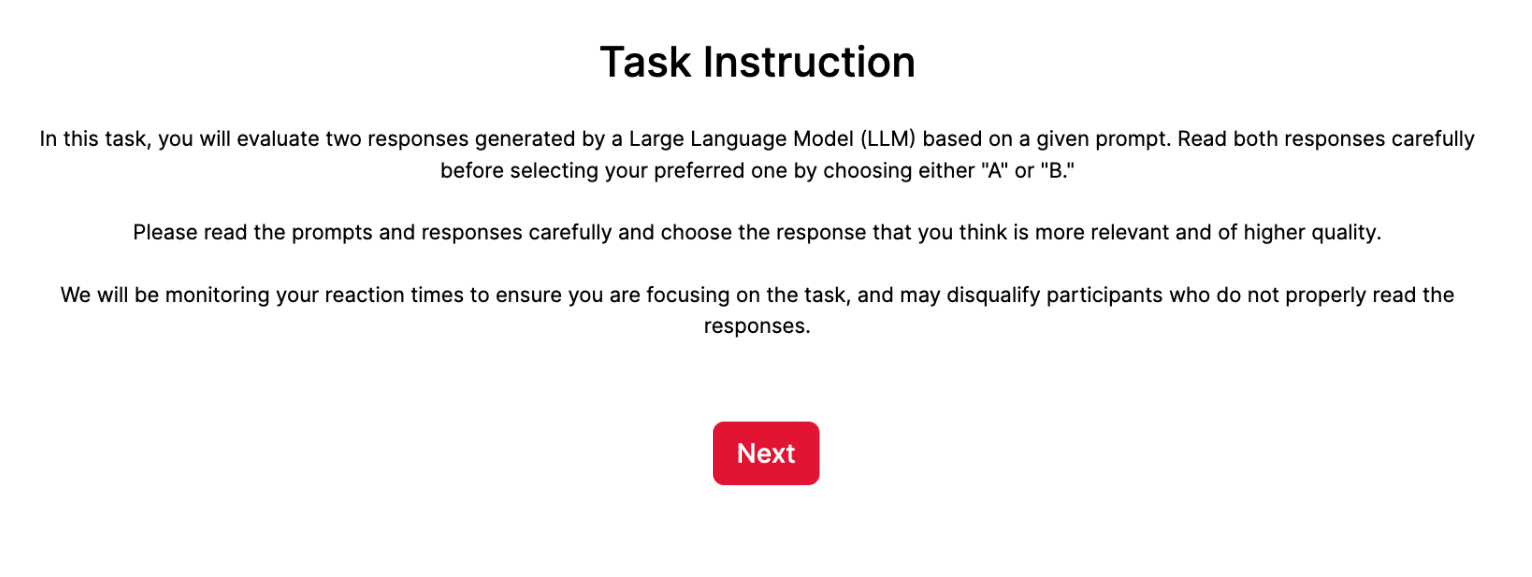}
    \caption{Instructions provided to participants for Experiment 3 (preference task). Participants compared two LLM-generated responses to the same prompt and selected their preferred response.}
    \label{fig:exp3-instruction}
\end{figure}

\begin{figure}[t]
    \centering
    \includegraphics[width=\linewidth]{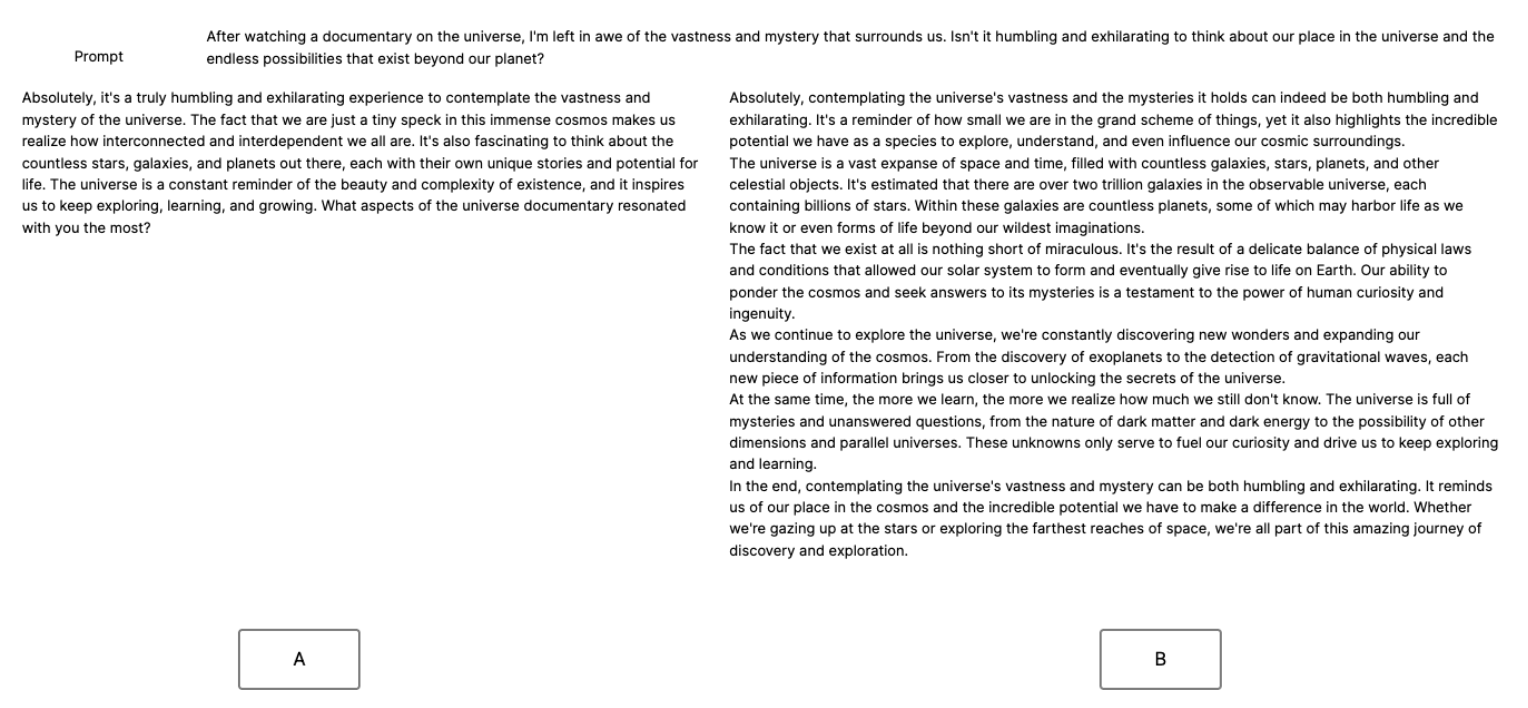}
    \caption{Example trial from Experiment 3, showing a prompt and two candidate responses. Participants chose which response they preferred based on perceived quality and relevance.}
    \label{fig:exp3-example}
\end{figure}

\begin{figure}[ht]
    \centering
    \includegraphics[width=\textwidth]{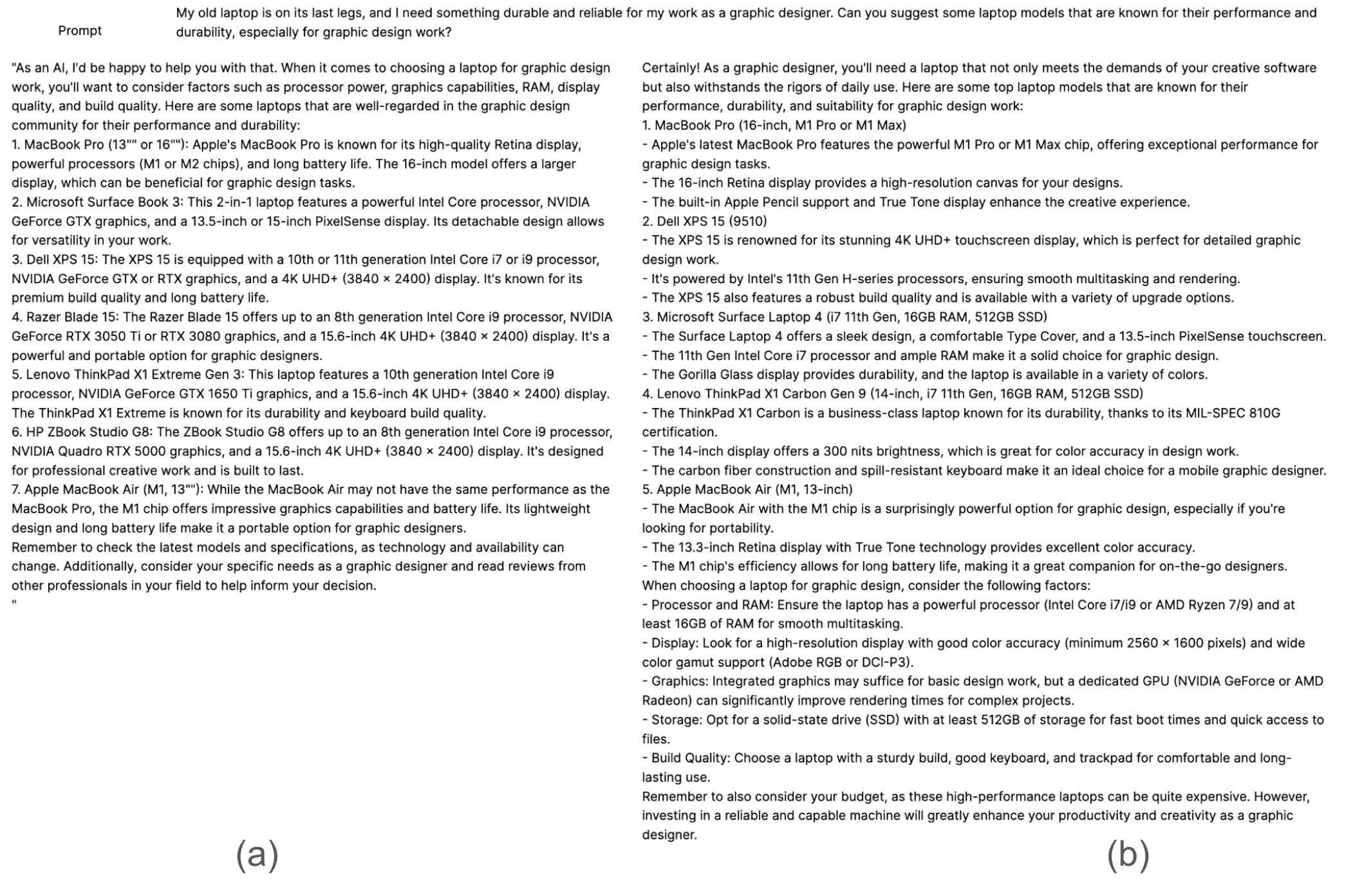}
    \caption{\textbf{Sample Information Seeking prompt}: Comparison between responses generated with an implicature-aware prompt (a) and a baseline prompt (b). This pair is included as an illustrative preferred example from the targeted comparison set; response (a) appears more tailored and contextually relevant.}
    \label{fig:information_seeking}
\end{figure}

\begin{figure}[ht]
    \centering
    \includegraphics[width=\textwidth]{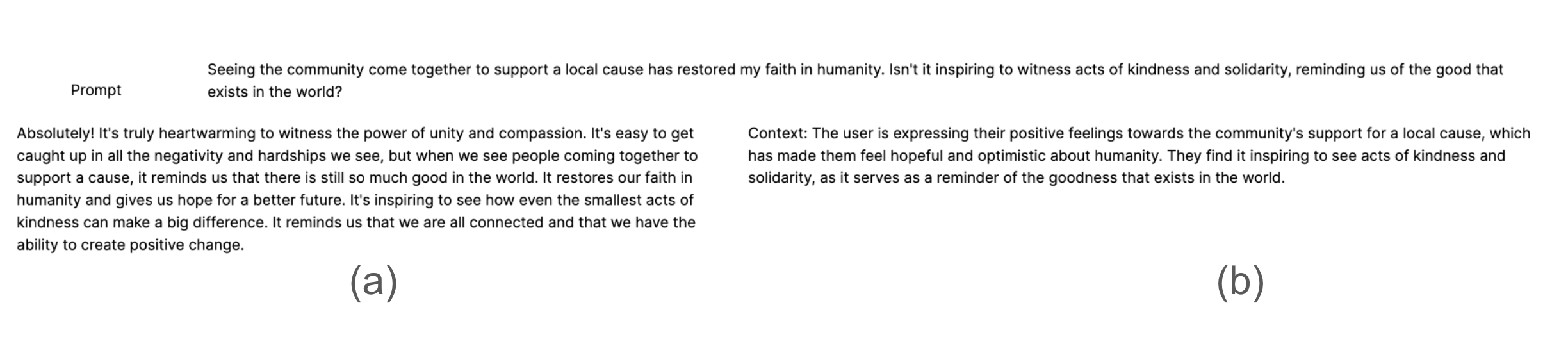}
    \caption{\textbf{Sample Expressive prompt:} Comparison of responses to an expressive prompt. This pair is included as an illustrative preferred example from the targeted comparison set; response (a) appears more empathetic and affect-sensitive, whereas response (b) focuses more on restating context.}
    \label{fig:expressive}
\end{figure}

\begin{figure}[ht]
    \centering
    \includegraphics[width=\textwidth]{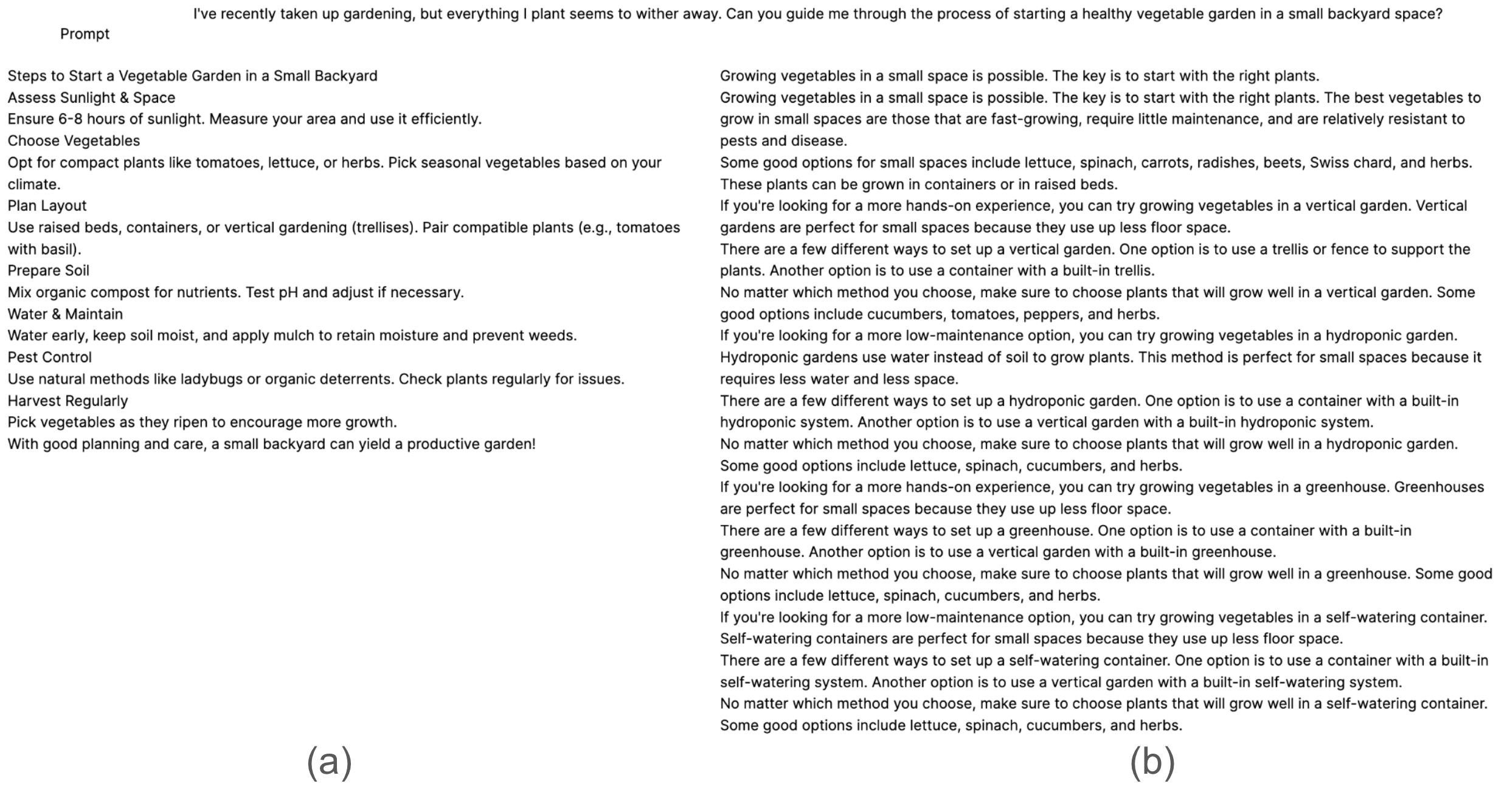}
    \caption{\textbf{Sample Direction Seeking prompt:} Comparison of responses to a direction-seeking prompt. This pair is included as an illustrative preferred example from the targeted comparison set; response (a) appears clearer and more structured, whereas response (b) is less targeted.}
    \label{fig:direction_seeking}
\end{figure}

\appendix

\section{Experiment 2 Model Reporting Details}

\begin{table}[ht]
\centering
\footnotesize
\begin{tabular}{llrr}
\toprule
\textbf{Outcome} & \textbf{Random Effect} & \textbf{Variance} & \textbf{SD} \\
\midrule
Relevance & prompt\_response & 0.091 & 0.301 \\
Relevance & participant & 0.225 & 0.475 \\
Relevance & Residual & 0.538 & 0.733 \\
Quality & prompt\_response & 0.100 & 0.317 \\
Quality & participant & 0.133 & 0.364 \\
Quality & Residual & 0.713 & 0.845 \\
\bottomrule
\end{tabular}
\caption{Variance components for the Experiment~2 linear mixed-effects models.}
\label{tab:app_e2_varcomp}
\end{table}

This appendix provides additional reporting details for the Experiment~2 mixed-effects analyses.  Table~\ref{tab:app_e2_varcomp} reports variance components for the relevance and quality models. Tables~\ref{tab:app_e2_rel_fixed} and~\ref{tab:app_e2_qual_fixed} report fixed-effect estimates with Wald 95\% confidence intervals. Because sum-to-zero contrasts were used, these coefficients are included for transparency and should be interpreted in conjunction with the Type~III tests reported in the main text.

\begin{longtable}{p{6.0cm}rrrrlp{3.5cm}}
\caption{Fixed effects for the Experiment~2 relevance model. Predictors were sum-coded; coefficients are reported for transparency and should be interpreted in conjunction with the Type~III tests in the main text.}\label{tab:app_e2_rel_fixed}\\
\toprule
\textbf{Term} & \textbf{Est.} & \textbf{SE} & \textbf{df} & \textbf{t} & \textbf{p} & \textbf{95\% CI}\\
\midrule
\endfirsthead
\toprule
\textbf{Term} & \textbf{Est.} & \textbf{SE} & \textbf{df} & \textbf{t} & \textbf{p} & \textbf{95\% CI}\\
\midrule
\endhead
\bottomrule
\endfoot
(Intercept) & 4.075 & 0.061 & 94.33 & 66.96 & $< .001$ & [3.954, 4.196]\\
intervention1 & -0.136 & 0.033 & 249.99 & -4.08 & $< .001$ & [-0.202, -0.070]\\
model1 & 0.186 & 0.064 & 248.09 & 2.89 & 0.004 & [0.060, 0.313]\\
model2 & 0.279 & 0.067 & 236.78 & 4.19 & $< .001$ & [0.148, 0.410]\\
model3 & -0.830 & 0.068 & 278.04 & -12.20 & $< .001$ & [-0.964, -0.696]\\
model4 & 0.204 & 0.065 & 249.07 & 3.12 & 0.002 & [0.075, 0.332]\\
class1 & 0.212 & 0.047 & 244.10 & 4.55 & $< .001$ & [0.120, 0.303]\\
class2 & 0.075 & 0.046 & 261.38 & 1.62 & 0.106 & [-0.016, 0.166]\\
intervention1 $\times$ model1 & -0.067 & 0.065 & 250.92 & -1.03 & 0.302 & [-0.195, 0.061]\\
intervention1 $\times$ model2 & 0.066 & 0.067 & 240.81 & 1.00 & 0.319 & [-0.065, 0.197]\\
intervention1 $\times$ model3 & -0.088 & 0.066 & 265.58 & -1.33 & 0.185 & [-0.219, 0.043]\\
intervention1 $\times$ model4 & 0.006 & 0.066 & 252.13 & 0.09 & 0.932 & [-0.124, 0.135]\\
intervention1 $\times$ class1 & 0.029 & 0.048 & 251.72 & 0.61 & 0.545 & [-0.065, 0.122]\\
intervention1 $\times$ class2 & 0.013 & 0.047 & 261.00 & 0.28 & 0.780 & [-0.079, 0.105]\\
model1 $\times$ class1 & -0.105 & 0.091 & 253.35 & -1.16 & 0.248 & [-0.284, 0.074]\\
model2 $\times$ class1 & -0.141 & 0.096 & 239.56 & -1.47 & 0.144 & [-0.329, 0.048]\\
model3 $\times$ class1 & 0.331 & 0.100 & 279.20 & 3.29 & 0.001 & [0.133, 0.529]\\
model4 $\times$ class1 & -0.028 & 0.090 & 235.15 & -0.32 & 0.752 & [-0.205, 0.149]\\
model1 $\times$ class2 & 0.024 & 0.091 & 258.74 & 0.26 & 0.795 & [-0.156, 0.204]\\
model2 $\times$ class2 & 0.061 & 0.090 & 232.33 & 0.67 & 0.501 & [-0.117, 0.239]\\
model3 $\times$ class2 & -0.208 & 0.092 & 271.85 & -2.25 & 0.025 & [-0.390, -0.026]\\
model4 $\times$ class2 & -0.016 & 0.092 & 255.58 & -0.17 & 0.864 & [-0.196, 0.165]\\
intervention1 $\times$ model1 $\times$ class1 & -0.048 & 0.094 & 272.22 & -0.51 & 0.610 & [-0.233, 0.137]\\
intervention1 $\times$ model2 $\times$ class1 & -0.035 & 0.095 & 233.31 & -0.36 & 0.716 & [-0.221, 0.152]\\
intervention1 $\times$ model3 $\times$ class1 & 0.129 & 0.101 & 277.82 & 1.28 & 0.202 & [-0.070, 0.327]\\
intervention1 $\times$ model4 $\times$ class1 & -0.062 & 0.090 & 231.85 & -0.68 & 0.497 & [-0.240, 0.117]\\
intervention1 $\times$ model1 $\times$ class2 & -0.129 & 0.094 & 270.51 & -1.37 & 0.172 & [-0.314, 0.056]\\
intervention1 $\times$ model2 $\times$ class2 & -0.026 & 0.091 & 233.51 & -0.28 & 0.778 & [-0.205, 0.154]\\
intervention1 $\times$ model3 $\times$ class2 & 0.168 & 0.092 & 266.63 & 1.82 & 0.070 & [-0.014, 0.349]\\
intervention1 $\times$ model4 $\times$ class2 & 0.062 & 0.092 & 253.40 & 0.67 & 0.502 & [-0.119, 0.243]\\
\end{longtable}

\begin{longtable}{p{6.0cm}rrrrlp{3.5cm}}
\caption{Fixed effects for the Experiment~2 quality model. Predictors were sum-coded; coefficients are reported for transparency and should be interpreted in conjunction with the Type~III tests in the main text.}\label{tab:app_e2_qual_fixed}\\
\toprule
\textbf{Term} & \textbf{Est.} & \textbf{SE} & \textbf{df} & \textbf{t} & \textbf{p} & \textbf{95\% CI}\\
\midrule
\endfirsthead
\toprule
\textbf{Term} & \textbf{Est.} & \textbf{SE} & \textbf{df} & \textbf{t} & \textbf{p} & \textbf{95\% CI}\\
\midrule
\endhead
\bottomrule
\endfoot
(Intercept) & 3.768 & 0.053 & 97.74 & 70.57 & $< .001$ & [3.662, 3.874]\\
intervention1 & -0.133 & 0.037 & 253.67 & -3.61 & $< .001$ & [-0.205, -0.061]\\
model1 & 0.249 & 0.071 & 250.66 & 3.48 & $< .001$ & [0.108, 0.390]\\
model2 & 0.365 & 0.073 & 238.25 & 4.98 & $< .001$ & [0.221, 0.510]\\
model3 & -1.194 & 0.075 & 283.83 & -15.89 & $< .001$ & [-1.342, -1.046]\\
model4 & 0.319 & 0.072 & 251.20 & 4.41 & $< .001$ & [0.176, 0.461]\\
class1 & 0.232 & 0.052 & 245.60 & 4.50 & $< .001$ & [0.131, 0.334]\\
class2 & 0.077 & 0.051 & 264.35 & 1.50 & 0.135 & [-0.024, 0.177]\\
intervention1 $\times$ model1 & 0.014 & 0.072 & 253.76 & 0.19 & 0.848 & [-0.128, 0.155]\\
intervention1 $\times$ model2 & -0.040 & 0.073 & 241.66 & -0.55 & 0.584 & [-0.185, 0.104]\\
intervention1 $\times$ model3 & -0.047 & 0.074 & 269.75 & -0.64 & 0.525 & [-0.192, 0.098]\\
intervention1 $\times$ model4 & -0.039 & 0.073 & 254.85 & -0.54 & 0.589 & [-0.183, 0.104]\\
intervention1 $\times$ class1 & 0.098 & 0.052 & 254.43 & 1.88 & 0.061 & [-0.005, 0.202]\\
intervention1 $\times$ class2 & -0.067 & 0.052 & 265.48 & -1.29 & 0.197 & [-0.169, 0.035]\\
model1 $\times$ class1 & -0.006 & 0.100 & 257.33 & -0.06 & 0.955 & [-0.204, 0.192]\\
model2 $\times$ class1 & -0.147 & 0.106 & 240.31 & -1.39 & 0.165 & [-0.356, 0.061]\\
model3 $\times$ class1 & 0.188 & 0.111 & 285.08 & 1.69 & 0.092 & [-0.031, 0.406]\\
model4 $\times$ class1 & -0.019 & 0.099 & 235.31 & -0.19 & 0.850 & [-0.215, 0.177]\\
model1 $\times$ class2 & -0.090 & 0.102 & 261.81 & -0.88 & 0.378 & [-0.290, 0.110]\\
model2 $\times$ class2 & 0.195 & 0.100 & 232.53 & 1.94 & 0.053 & [-0.003, 0.392]\\
model3 $\times$ class2 & -0.281 & 0.102 & 276.91 & -2.74 & 0.007 & [-0.483, -0.079]\\
model4 $\times$ class2 & 0.017 & 0.102 & 257.83 & 0.17 & 0.867 & [-0.184, 0.218]\\
intervention1 $\times$ model1 $\times$ class1 & -0.047 & 0.103 & 276.49 & -0.46 & 0.646 & [-0.250, 0.156]\\
intervention1 $\times$ model2 $\times$ class1 & 0.031 & 0.105 & 233.97 & 0.29 & 0.772 & [-0.176, 0.237]\\
intervention1 $\times$ model3 $\times$ class1 & 0.223 & 0.111 & 284.19 & 2.01 & 0.046 & [0.004, 0.442]\\
intervention1 $\times$ model4 $\times$ class1 & -0.184 & 0.100 & 233.24 & -1.84 & 0.067 & [-0.381, 0.013]\\
intervention1 $\times$ model1 $\times$ class2 & -0.098 & 0.104 & 275.49 & -0.94 & 0.346 & [-0.302, 0.106]\\
intervention1 $\times$ model2 $\times$ class2 & 0.000 & 0.101 & 234.33 & 0.00 & 0.998 & [-0.198, 0.199]\\
intervention1 $\times$ model3 $\times$ class2 & 0.084 & 0.102 & 272.27 & 0.82 & 0.413 & [-0.118, 0.285]\\
intervention1 $\times$ model4 $\times$ class2 & 0.129 & 0.102 & 256.41 & 1.26 & 0.210 & [-0.073, 0.330]\\
\end{longtable}

\end{document}